\documentclass[10pt,twocolumn,letterpaper]{article}

\usepackage[pagenumbers]{cvpr} 

\usepackage{graphicx}
\usepackage{amsmath}
\usepackage{amssymb}
\usepackage{booktabs, comment}

\usepackage{url}
\usepackage[utf8]{inputenc} 
\usepackage[T1]{fontenc}    
\usepackage{subcaption}
\usepackage[normalem]{ulem} 
\usepackage[export]{adjustbox}
\usepackage{amsfonts}       
\usepackage{nicefrac}       
\usepackage{microtype}      
\usepackage{gensymb}
\usepackage{tabularx}
\usepackage{siunitx}
\usepackage{enumitem}
\usepackage{lineno}
\usepackage{pifont}
\usepackage{upgreek}
\usepackage{bbm}
\usepackage{scalerel}

\usepackage[font=small,labelfont=bf]{caption}

\usepackage{stfloats}

\newcommand{\RN}[1]{%
  \textup{\uppercase\expandafter{\romannumeral#1}}%
}

\definecolor{navy}{rgb}{0.7, 0.1, 0.7}
\definecolor{burgundy}{RGB}{144,0,32}
\definecolor{green2}{RGB}{0,160,0}

\newcommand{\KGnote}[1]{{\color{magenta}{\bf KG: }#1}} 
\newcommand{\KG}[1]{{\color{blue}#1}} 

\newcommand{\ZAnote}[1]{{\color{teal}{\bf ZA: }#1}} 
\newcommand{\ZA}[1]{{\color{cyan}#1}} 

\newcommand{\TNnote}[1]{{\color{ForestGreen}{\bf TN: }#1}} 

\newcommand{\SMnote}[1]{{\color{red}{\bf SM: }#1}} 
\newcommand{\SM}[1]{{\color{burgundy}#1}} 

\renewcommand{\KG}[1]{{\color{black}#1}} 
\renewcommand{\ZA}[1]{{\color{black}#1}} 
\renewcommand{\SM}[1]{{\color{black}#1}} 

\renewcommand{\KGnote}[1]{{\color{magenta}}} 
\renewcommand{\ZAnote}[1]{{\color{teal}}} 
\renewcommand{\TNnote}[1]{{\color{ForestGreen}}} 
\renewcommand{\SMnote}[1]{{\color{red}}} 

\newcommand{\cider}{CIDEr}

\newcommand{\meteor}{METEOR}
\newcommand{\iou}{IoU}

\definecolor{cvprblue}{rgb}{0.21,0.49,0.74}
\usepackage[pagebackref,breaklinks,colorlinks,citecolor=cvprblue]{hyperref}

\usepackage[capitalize]{cleveref}
\crefname{section}{Sec.}{Secs.}
\Crefname{section}{Section}{Sections}
\Crefname{table}{Table}{Tables}
\crefname{table}{Tab.}{Tabs.}

\def\paperID{xxxxx} 
\def\confName{CVPR}
\def\confYear{2025}

\usepackage{fancyhdr}

\fancypagestyle{firstpage}{%
  \fancyhf{}
  \chead{\small In Proceedings of IEEE/CVF Conference on Computer Vision and Pattern Recognition (CVPR), 2025}
}

\title{Which Viewpoint Shows it Best? Language 
for Weakly Supervising \\View Selection in Multi-view 
Instructional
Videos}

\author{%
Sagnik Majumder$^{1,2}$ \hspace{3mm} Tushar Nagarajan$^{2}$ \hspace{3mm} Ziad Al-Halah$^{3}$ \hspace{3mm} Reina Pradhan$^{1}$ \hspace{3mm} Kristen Grauman$^{1,2}$\\
$^1$UT Austin \hspace{3mm} $^2$FAIR, Meta \hspace{3mm} $^3$University of Utah 
}

\begin{document}
\maketitle

\thispagestyle{firstpage}

\begin{abstract}
Given a multi-view video, which viewpoint is most informative for a human observer?  Existing methods rely on heuristics or expensive ``best-view" supervision to answer this question, limiting their applicability.  We propose a weakly supervised approach that leverages language accompanying an instructional multi-view video as a means to recover its most informative viewpoint(s).  Our key hypothesis is that the more accurately an individual view can predict a view-agnostic text summary, the more informative it is.  To put this into action, we propose \textsc{LangView}, a framework that uses the relative accuracy of view-dependent caption predictions as a proxy for best view pseudo-labels.  Then, those pseudo-labels are used to train a view selector, together with an auxiliary camera pose predictor that enhances view-sensitivity.  During inference, our model takes as input only a multi-view video---no language or camera poses---and returns the best viewpoint to watch at each timestep.  On two challenging datasets comprised of diverse multi-camera setups and how-to activities, our model consistently outperforms state-of-the-art baselines, both with quantitative metrics and human evaluation. Project: \url{https://vision.cs.utexas.edu/projects/which-view-shows-it-best}.
\end{abstract}

\section{Introduction}\label{sec:introduction}

\begin{figure}[!ht]
    \centering
    \includegraphics[width=0.8\linewidth]{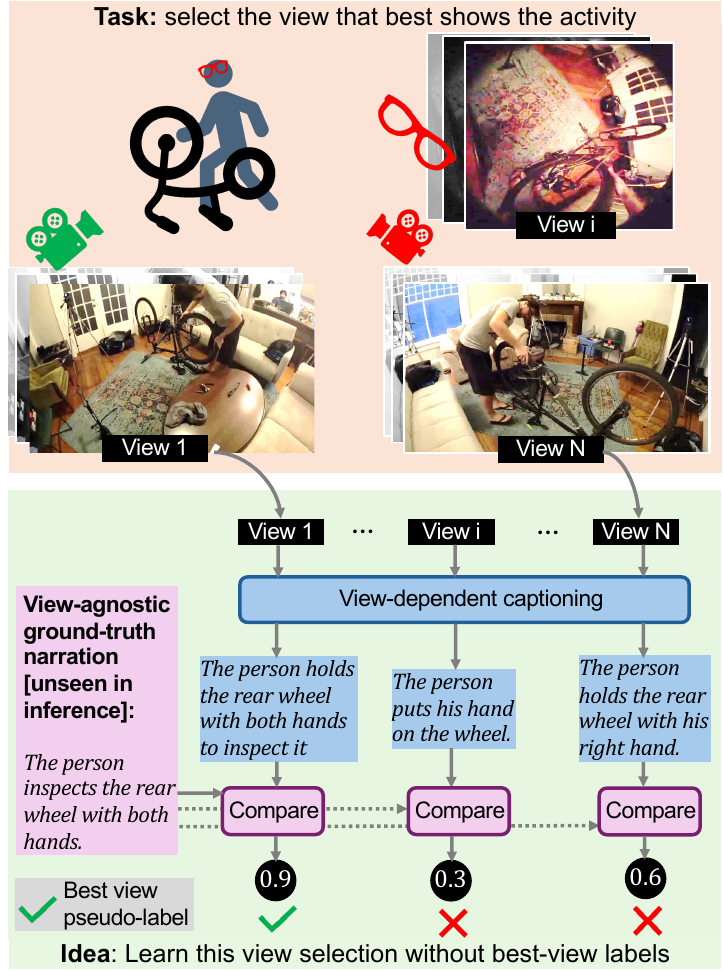}
    \caption{
     \SM{\textsc{LangView} idea:} 
    given multi-view instructional videos, we aim to learn a view selection model that can identify the best view for \KG{seeing} how to perform the activity shown in the videos, in the \emph{absence of best view labels}. To achieve this, we compare each estimated view-dependent caption to the view-agnostic ground-truth video narration of the human activity, and use their respective accuracies as a proxy for view quality.  These quality scores then serve as pseudo-labels for learning to select the most informative view. In this example, the \KG{1st} view most clearly shows all entities involved in the activity---the wheel and the person's hands, and how they interact---and hence, produces a caption that best matches the ground-truth, making it a positive pseudo-label for view selection.
    }
   \vspace{-0.5cm}
    \label{fig:intro}
\end{figure}

Videos are an essential vehicle for communicating how to perform a new skill, as evidenced by the millions of ``how-to" videos online, for everything from frosting a cake to perfecting a basketball layup.  The more intricate the task, however, the more important the \emph{viewpoint} used to film the instructional video.   For example, a close-up view of the hands is desirable when a knitter shows how to add stitches of yarn to a needle, or when a rock-climber demonstrates a particular hold---whereas a view from afar may be preferable when the knitter shows the knitted sweater being worn, or the climber shows their selected path up the wall.  In general, the information available in any given viewpoint of an activity varies.   Not all views are created equal.

Shooting a video with multiple cameras 
provides a holistic view of the activity taking place in a scene, by capturing it from different locations and angles,
and \emph{multi-view video} is developing as a new frontier in computer vision research~\cite{sigurdsson2018actor, tang2019multi, mildenhall2020nerf, grauman2023ego, huang2024egoexolearn}, especially in instructional settings~\cite{miech2019howto100m, grauman2023ego}.   
However, multi-view videos are generally not suitable for direct human consumption~\cite{feaf4658d920469e95d5720e873bdb97}: digesting multiple views at once imposes a high cognitive burden.  Thus, in practice, the status quo is to orchestrate view selection in how-to videos manually with either active camera work or post-production video editing tools, which is time consuming and tedious.

What if instead a vision model learned to automatically perform \emph{view selection}, at every time step deciding which camera from the multi-view video to adopt?  View selection has traditionally been studied in the context of automatic cinematography for specialized domains, e.g., 360$^{\circ}$ panoramas~\cite{su2016pano2vid, Lee2018AMN}, sports clips~\cite{chen2019learning,hu2017deep}, virtual environments~\cite{10.5555/3022508.3022511, 10.1145/237170.237259, 10.1145/3596711.3596786, 10.1145/2788539.2788549}, or lecture videos~\cite{871033, 10.1145/1324287.1324293, 10.1109/TMM.2005.854388}. 
Aside from their specialized domains, existing work is limited by relying on hand-coded heuristics~\cite{10.5555/3022508.3022511, 10.1145/237170.237259, 10.1145/2601097.2601198} or assuming access to manual labels indicating the favored views for training~\cite{su2016pano2vid, 
Lee2018AMN, hu2017deep,
Xiong_2018_ECCV, chou2017self,
chen2019learning}.
\KG{Such labels are expensive and quite special purpose.} 

Conscious of these shortcomings, we propose to learn 
view selection 
in multi-view instructional videos in the \emph{absence of 
best view labels}.
Towards that goal, we hypothesize that 
\emph{view-agnostic} natural language descriptions of 
the activity shown in the videos~\cite{grauman2023ego, jia2020LEMMA, huang2024egoexolearn}---commonly referred to as ``narrations"\footnote{
Narrations in multi-view datasets~\cite{grauman2023ego,jia2020LEMMA,huang2024egoexolearn} are produced by human annotators who watch all views and write down a view-independent description of how the activity is performed, and in the wild they correspond to the ``how-to" descriptions spoken by a person demonstrating a task~\cite{miech2019howto100m}.},
can act as a source of 
weak 
supervision. 
Specifically,  
our core idea is
that for any multi-view video clip, the viewpoint that is 
most
predictive of such a narration is likely to be 
the most informative of the activity, and hence, can be pseudo-labeled as the best view for training a view selector.
For example, given a multi-view video of a person repairing a bike (Figure~\ref{fig:intro}), independent captions on each view will emphasize different visible components of the scene (the wheel, the person's hands, other objects in the scene, etc.); the caption most aligned with the view-agnostic narration 
\SM{``\emph{the person removes the rear wheel with both hands}"}
indicates which view is most informative for the \emph{whole} activity content in that clip.  \KG{Unlike explicit best-view labels, the vision-language annotations that fuel today's captioners are open-world, versatile, and widely available.}

To validate our hypothesis, we design a novel framework called \textsc{LangView}, which is
composed of two key elements: a best view pseudo-labeler, and a best view selector. The 
pseudo-labeler automatically generates best view pseudo-labels for a multi-view video during training, by using 
off-the-shelf video captioners\SM{~\cite{zhang2023video, li2023mvbench}} to 
score and rank views on the basis of how well the predicted
narration from a view matches the view-agnostic
ground-truth narration. 
The 
selector
takes a multi-view video as input, and
predicts 
the best view labels.  During training, the selector also solves an auxiliary task of predicting the relative camera pose between different views, to increase its view-sensitivity and improve its selection accuracy. 
At inference, our model requires as input only a multi-view video, but no language or camera poses.

We evaluate
\textsc{LangView}
using two challenging multi-view 
instructional video
datasets encompassing diverse activity scenarios and multi-camera
setups, 
Ego-Exo4D~\cite{grauman2023ego} and LEMMA~\cite{jia2020LEMMA}. On both, our model outperforms multiple baselines and state-of-the-art methods for view selection 
on several automatic and human evaluation metrics. 
More broadly, our work offers a novel way for language to elicit the information content of video. 

\section{Related work}\label{sec:related}

\paragraph{Temporal video summarization.}
Temporal video summarization~\cite{panda2017collaborative, 10.1007/978-3-030-58589-1_16, park2020sumgraph, 
bhattacharya2021highlightme, narasimhan2022tl,
He_2023_CVPR}
creates a synopsis of a long video by identifying and stitching together 
its most representative clips. 
Early  unsupervised methods use hand-crafted features, 
optimizing
for metrics 
like
object saliency~\cite{10.1109/CVPR.2013.350}
and motion attention~\cite{1238320}, while
recent efforts~\cite{7298928, NIPS2014_0eec27c4, li2018local, 
lee2015predicting} 
use 
labeled data~\cite{7299154, 10.1007/978-3-319-10584-0_33} for supervised training.
To mitigate the supervision cost, 
recent work
leverage web priors~\cite{6619192, 6909891, c4799d56bda249b18e47f112d550c485} to target settings with insufficient or unreliable annotations~\cite{6619192, yang2015unsupervised, 
7299154, 8237657} or unpaired data~\cite{rochan2019video, Ye_2021_ICCV, narasimhan2022tl}. 
Different from all of the above, we tackle the problem of label scarcity in the context of \emph{view selection}
in multi-view 
videos, a distinct task from video summarization.  
In multi-video summarization, the input videos are either  
captured with multi-view cameras in indoor 
surveillance~\cite{panda2017multi, 5482155} 
or street~\cite{elfeki2022multi, 8824208, 5482155} settings, or grouped 
using
shared visual 
concepts~\cite{panda2017collaborative, 8265353, 10.1145/3343031.3350938, 8546119, 7298981}.
Such
methods can  
select multiple views at the same time, 
and hence, are not applicable 
in our setting.  
Unlike all video summarization methods, where the goal is to create a \emph{sparse} temporal summary of one or more videos, our task requires choosing the appropriate camera view at 
\emph{each} time step (clip).
\vspace{-0.25cm}

\paragraph{Automatic cinematography.}
Automatic cinematography involves automatically choosing the best camera angles, positions, and zoom levels for human consumption of a 
video scene.
Existing work explores camera control in virtual environments~\cite{10.5555/3022508.3022511, 10.1145/237170.237259, 10.1145/3596711.3596786, 10.1145/2788539.2788549}, or addresses a narrow domain like lecture videos~\cite{871033, 10.1145/1324287.1324293, 10.1109/TMM.2005.854388,10.1145/569005.569007,10.1145/1180639.1180851}, social settings with multiple egocentric cameras~\cite{10.1145/2601097.2601198}, 
or panoramic 360$^\circ$ input videos~\cite{su2016pano2vid,Su2017Making3V,Lee2018AMN,chou2017self,Xiong_2018_ECCV}. 
Prior learning-based methods
require
manual labels
to guide 
selection~\cite{su2016pano2vid,Su2017Making3V, Lee2018AMN, hu2017deep, chen2019learning}.  
In 
contrast, we
show how to 
train a view selector \emph{without} \ZA{manual} labels, by exploiting the 
accuracy of predicted narrations from different views as a proxy for view quality. In
continued complementary research, we explore how to leverage single-view instructional videos during training~\cite{majumder2024switch}
\vspace{-0.25cm}

\paragraph{Active next view selection.}
Active next view selection~\cite{5968} requires an embodied agent to 
smartly control its camera for solving tasks like object recognition~\cite{8367872, jayaraman2016look, ammirato2017dataset, cheng2018geometry, ramakrishnan2018sidekick, ramakrishnan2019emergence, Du_2023_ICCV}, reconstruction~\cite{jayaraman2018learning, ramakrishnan2018sidekick, ramakrishnan2019emergence, seifi2021glimpse, Jha_2023_WACV}, and 
semantic segmentation~\cite{seifi2020attend, seifi2021glimpse}
within a 
time budget.
Whereas such setups
require
moving the camera to capture a better view  
for the agent performing its task, 
our goal is to select
the best view from 
multiple cameras recording simultaneously to facilitate
human viewing. 
\vspace{-0.25cm}

\paragraph{Captioning for 
weak
supervision.}
Prior work 
studies
using ground-truth~\cite{desai2021virtex, tschannen2024image, yu2022coca} or predicted~\cite{doveh2024dense, yang2023alip} captions for
weakly supervising
action recognition~\cite{Gupta_Mooney_2010}, object 
detection~\cite{sun2015automatic, Ye_2019_ICCV, zareian2021open, 9879567, liu2023grounding, kong2024hyperbolic}, 
semantic segmentation~\cite{xu2023open, vs2023mask, Wu_2023_ICCV}, 
and 
visual question answering~\cite{banerjee2020weaqa, yang2022learning, guo2022images}. 
On the contrary, we 
\SM{tackle view selection, a distinct task, by using the quality of predicted~\cite{zhang2023video,li2023mvbench} captions (narrations) as weak supervision.}

\section{Approach}\label{sec:approach}

\begin{figure*}[t]
    \centering
    \begin{subfigure}[b]{0.59\linewidth}
    \centering
    \includegraphics[width=\linewidth]{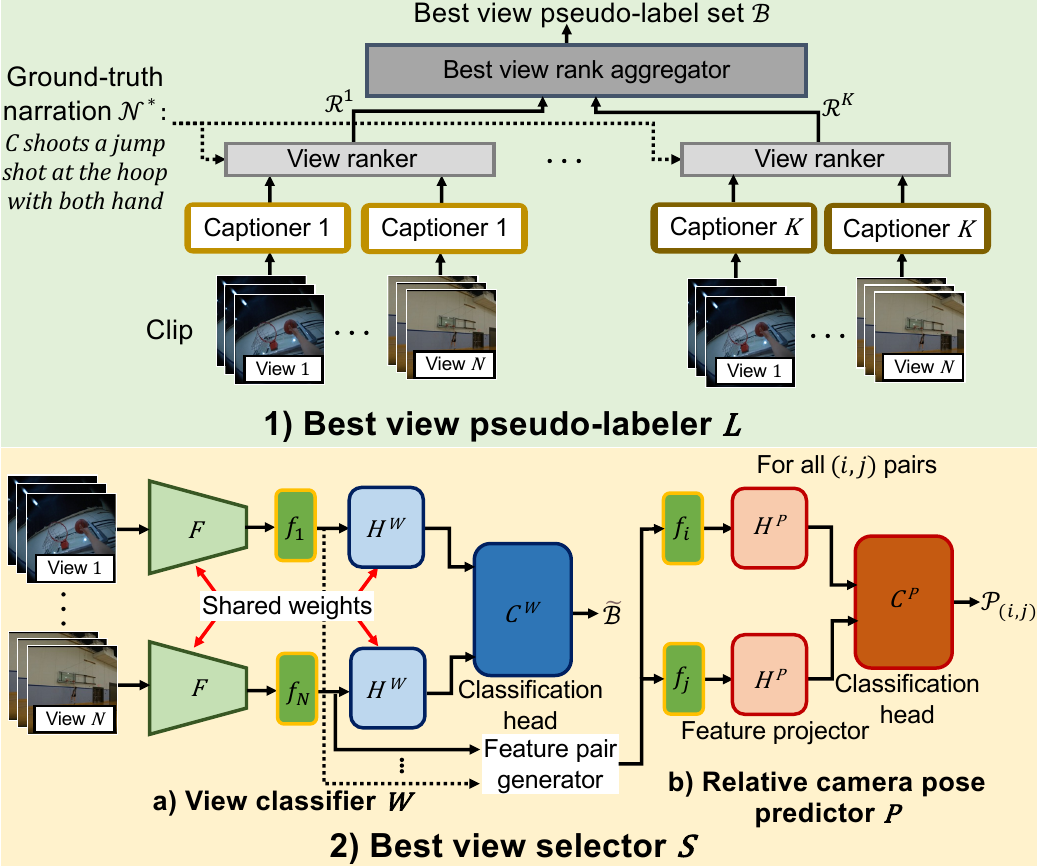}
    \caption{Language-guided view selection training framework}
    \label{fig:model}
    \end{subfigure}\hfill
    \begin{subfigure}[b]{0.385\linewidth}
    \centering
    \includegraphics[width=\linewidth]{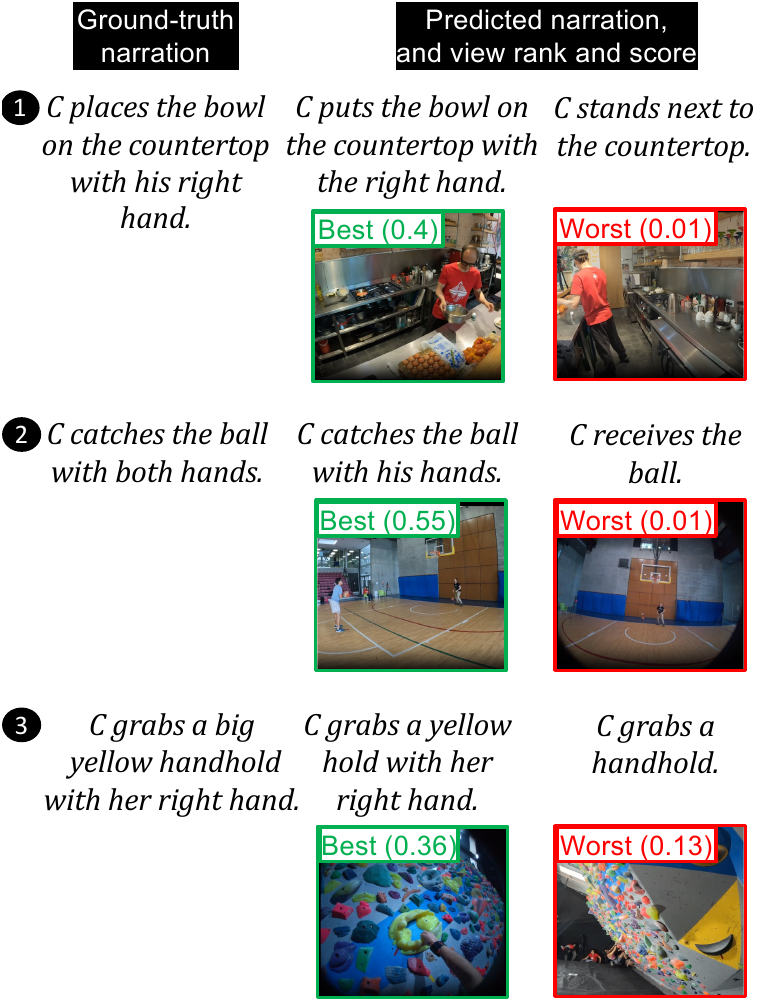}
    \caption{Ground-truth narration and pseudo-labeler output samples 
    }
    \label{fig:pseudolabel_examples}
    \end{subfigure}\hfill
    \vspace*{-0.3cm}
\caption{\textbf{(a)} Our model uses language guidance to train a 
view-selector for multi-view instructional videos, such that the chosen views help best understand the shown activity.
To do so, we first generate best view pseudo-labels during training by leveraging clip narrations, where each narration is a \emph{view-agnostic} and detailed description of the activity.
Specifically, given a training clip, we use off-the-shelf video captioners to predict 
\KG{a caption}
per view, score the views by comparing their 
\KG{captions} to the ground-truth \KG{narration},
and finally rank the views to generate a best view pseudo-label for the clip. Given 
the 
multi-view 
clip, 
our view classifier (bottom-left) encodes 
it with
a visual encoder, 
and predicts 
a pseudo-label
estimate. We also solve
an auxiliary task of relative camera pose prediction (bottom-right)
that
increases the view sensitivity of the classifier. \textbf{(b)} Examples of predicted narrations, and the ranks and scores of the views 
per
our pseudo-labeler, 
shown alongside  
ground-truth view-agnostic narrations.  “C" refers to the person who is performing the activity. \KG{Note that at inference time, there is no ground truth narration, just the video input.}
}
\label{fig:model_n_pseudoLabelExamples}
\vspace{-0.35cm}
\end{figure*}

Our goal is to train a model to identify 
a sequence of best views 
for watching a 
multi-view instructional video, such that the identified views
are most informative of the activity in the video. 
Importantly, we aim to do this in the \emph{absence of 
best view labels.}
We first formally define our task (Sec.~\ref{sec:task}), and then decompose 
it
into three key questions: 
(1) 
how to source the best view pseudo-labels to train our model (Sec.~\ref{sec:labeler}), 
(2) how to model multi-view videos to discriminate between visually similar views 
when identifying the best one (Sec.~\ref{sec:selector}),
and
(3) how to train the model (Sec.~\ref{sec:training}).

\subsection{View selection task}\label{sec:task}

Given a set of 
instructional videos 
with multiple camera views, the goal is to 
automatically predict view\SM{s} 
that most comprehensively capture the fine-grained details---minute aspects of the actions and objects involved, precise body movements---of the human activity, and are, consequently, likely useful for 
skill learning. 
Critically, we aim to achieve this without any manually provided 
best-view labels,
but by instead using natural language narrations of the videos 
as a source of weak supervision. 

Let $V$ be 
an instructional video recorded with 
multiple cameras (Fig.~\ref{fig:intro}).
The video $V$ consists of $M$ clips, such that $V = \big[\mathcal{V}_{1}, ..., \mathcal{V}_{M}\big]$. Each clip $\mathcal{V}_{m}$ has $N$ RGB image streams, one from each camera/viewpoint, such that 
$\mathcal{V}_m = \big\{\mathcal{V}_{m, 1}, ...,\mathcal{V}_{m, N} \big\}$. 
Our goal is to select the best view 
$\mathcal{B}^{*}_m$
for each clip $\mathcal{V}_m$ to create an output video that is ideal for  
understanding
the activity in the video.
Therefore, we aim to find 
$\mathcal{B}^{*} = [\mathcal{B}^{*}_1, \ldots, \mathcal{B}^{*}_M]$,
where 
$\mathcal{B}^{*}_m \in \{\mathcal{V}_{m,i}\}_{i=1}^N$.

Rather than assume any best-view labels on the training clips $\mathcal{V}_m$, we turn instead to  \emph{narrations}---human-provided  descriptions of the activity in the video.  Narrations are common in instructional videos~\cite{zhou2018towards, miech2019howto100m, grauman2022ego4d, grauman2023ego, jia2020LEMMA,Damen2018EPICKITCHENS},
and for recent multi-view datasets~\cite{grauman2023ego, jia2020LEMMA} they are specifically gathered in a \emph{view-agnostic} manner:
human annotators 
watch a collage of \emph{all} views of a clip and write down a holistic description 
of the person’s actions and the objects involved. 
See Fig.~\ref{fig:model_n_pseudoLabelExamples} (center). While the narrations themselves are a form of
annotation, they are more widely available, versatile, and
scalable than specialized best-view labels, and hence provide
a compelling source of weak supervision.

Each clip thus has a ground-truth view-agnostic narration  $\mathcal{N}_{m}^{*}$. 
We stress that each $\mathcal{N}_{m}^{*}$ captures the activity as viewed from any and all angles; it is view-independent.
The descriptions include details about actions taken by the camera wearer
to the activity, and
as well as relevant events from the environment and important objects.  
We aim to use these narrations to train a 
model $\mathcal{F}$ that, given the video $V$, predicts a 
viewpoint sequence
i.e., 
$\mathcal{F}(V) = \mathcal{B}^{*}$. 
We stress that the narrations are available only 
for training videos, not at test time. 

\subsection{Sourcing best view pseudo-labels for training} 
\label{sec:captions}
Next, we describe our framework (Fig.~\ref{fig:model}) for tackling this 
\emph{language-guided}
task. 
We first source best-view pseudo-labels for 
training our view selector. 
We hypothesize that the relative quality of 
predicted narrations
from different views indicates how accurately each view captures the fine details of the activity,
and we show how this view-dependent quality of  predicted narrations can 
be used to 
train
a
view selector.
For example, consider a video showing how to 
fix a bicycle (Fig~\ref{fig:intro}). Some camera angles may have the person's body or the bike blocking the view, making it hard to see what is happening. As a result, the 
captioner
cannot properly 
describe 
the 
activity using such views, indicating that these views are of poor quality. 
In contrast, a view that clearly shows the hands, 
bike parts, tools, etc.,
will allow the 
captioner
to accurately describe the activity, making it 
a more informative view.

\paragraph{Best view pseudo-labeler \SM{$L$}.}\label{sec:labeler}

We devise our pseudo-labeler $L$ to generate the best-view pseudo-label for a training clip $\mathcal{V}_m$ by first 
predicting the narration for
each view separately, and then scoring the views by comparing their 
predicted narrations
to the \emph{view-agnostic} ground-truth narration $\mathcal{N}_m^{*}$.\footnote{For simplicity, we omit the clip index $m$ from subscript, henceforth.}
We aggregate results over multiple independent captioners in order to bolster robustness, essentially smoothing over their outputs.
To this end, we 
\SM{use}
$K$ off-the-shelf video captioning models\SM{:}
1) Video-Llama~\cite{zhang2023video} with Llama2~\cite{touvron2023llama2} LLM decoder, 2) Video-Llama~\cite{zhang2023video} with Vicuna~\cite{touvron2023llama2} LLM decoder, and 3) VideoChat2~\cite{li2023mvbench}.
See Fig.~\ref{fig:model} top.  

In particular, we 
predict the narrations for the $N$ views separately using each captioner, where $\mathcal{N}^k = \{\mathcal{N}_{1}^k, \ldots, \mathcal{N}_{N}^k\}$ denotes the 
predicted narrations
from the $k^\emph{th}$ captioner. Next, we pass each set $\mathcal{N}^k$ to a view ranker, which scores its 
narrations
by comparing them to the ground-truth narration $\mathcal{N}^*$ using a standard captioning metric~\cite{Vedantam2014CIDErCI, papineni-etal-2002-bleu, banerjee-lavie-2005-meteor}, and computes a set of ranks $\mathcal{R}^k$ for the corresponding views. 
Finally, a best view 
rank aggregator is used to reach an agreement across all captioners.
The 
rank aggregator extracts the consensus: 
it takes as input all rank sets $\{\mathcal{R}^1, \ldots, \mathcal{R}^K\}$, finds the views\SM{---there could be  multiple such views for a captioner if the estimated narration from more than one view produces the same highest score---}ranked the highest \SM{within each individual captioner,}
and uses 
\SM{\emph{all} views that are top-ranked across all captioners}
to build a best view pseudo-label set 
$\mathcal{B}$.

Essentially, our 
pseudo-labeler 
uses
multiple strong
captioners
to rank views 
based on the accuracy of their predicted 
narrations,
and achieves 
consensus
on the top ranked views,
thereby automatically 
producing
high-quality best view pseudo-labels.
See Fig.~\ref{fig:pseudolabel_examples} and Supp.~for examples.

\subsection{Best view selector \SM{$S$}}\label{sec:selector} 

We use the pseudo-label set 
$\mathcal{B}$
from Sec.~\ref{sec:labeler} to train our best view selector $S$, which must reason across all views to identify the best one \footnote{Note that it is not possible to simply apply the pseudo-labeler at inference time, since it requires access to ground-truth narrations.}. 
Simply using a view classifier as the selector is not enough in our setting, as our captioning-based labels can \SM{sometimes} be insufficiently discriminative, \ie
\SM{, multiple}
views 
might be pseudo-labeled as the best view due to the very nature of 
the off-the-shelf video captioners' training. They
\KG{were originally} designed to learn features that are predictive of the narration 
regardless of the view,
and consequently, can end up collapsing all views into  similar representations that are unable to capture the important nuances between 
them.

To tackle this,
we design 
a view selector 
composed of
1) a view 
classifier
$W$, and 2) a relative camera pose predictor $P$. 
\KG{During training,} while our view 
classifier
(Fig.~\ref{fig:model} bottom-left) tries to identify the best viewpoint given our pseudo-labels, the pose predictor (Fig.~\ref{fig:model} bottom-right) 
simultaneously \SM{acts as a regularizer and} mitigates the effect of spurious \SM{pseudo-}labels by solving 
an auxiliary task of relative camera pose prediction for each pair of viewpoints. This 
ensures that the features learned by our view 
classifier
remain sensitive to viewpoint changes\SM{, and our model does not overfit to the pseudo-labels during training}, thereby improving the quality of 
view selection, \KG{as we will see in results}.

\paragraph{View classifier \SM{$W$}.}
Our view 
classifier
$W$ consists of a visual encoder $F$ with a TimeSformer~\cite{bertasius2021space, pramanick2023egovlpv2} architecture,
which encodes each view $n$ into a set of visual features $f_n$ that spatially correspond with frame patches in the input view,
\SM{where $f_n = F(\mathcal{V}_n)$}. 
Owing to its patch-level nature and the end-to-end training (cf.~Sec.~\ref{sec:training}) 
of the 
classifier
using our 
pseudo-labels,
$f_n$ provides fine-grained cues about the human activity---what parts of it are visible in a viewpoint, and how the dynamic elements (\eg moving objects, body parts) evolve over time,  
thereby facilitating high-quality view selection. 
Next, the model uses a projector $H^W$ to embed $f_n$ into a lower-dimensional feature $h_n$ providing a higher-level representation of a view's ability to capture activity 
details.
\SM{Formally, $h_n = H^{W}(f_n)$.}
Finally, we concatenate $h_n$ from all views, and feed them to a 
classification head
$C^W$.
The 
classification head
compares these representations across views 
and outputs its best view estimate 
$\tilde{\mathcal{B}}$,
\SM{such that $\tilde{\mathcal{B}} = C^W([h_1, \ldots, h_N])$.}

\paragraph{Relative camera pose predictor \SM{$P$}.}
Our relative camera pose predictor $P$ uses the 
view 
classifier's
visual features to predict the relative camera pose for all
view pairs. 
Specifically, we formulate the 
pose prediction as a classification task~\cite{chen2021wide, nagarajan2023egoenv}. 
Given the ground-truth relative camera pose $P^*_{(i, j)}$ for an arbitrary pair of viewpoints $(i, j)$, we 
discretize the angles of its direction of displacement and rotation matrix 
using bins of a fixed size. 
This formulation 
\SM{ensures that our pose prediction task is tractable, and}
helps our model learn view-dependent visual features that improve task performance.
That is because it requires predicting the rough direction of one camera center relative to another
instead of the exact relative displacement,
which 
\SM{is ill-posed}
due to unknown object sizes.

To perform this classification for a viewpoint pair $(i, j)$, our 
pose predictor $P$ uses the fine-grained visual features $f_i$ and $f_j$ produced by our view 
classifier
for viewpoints $i$ and $j$, and embeds them into more abstract representations $h^P_i$ and $h^P_j$ by using a feature projector $H^P$,
\SM{such that $[h^P_i, h^P_j] = [H^P(f_i), H^P(f_j)]$.}
Whereas the fine-grained features $f$ help learn patch-level correspondences~\cite{bertasius2021space, pramanick2023egovlpv2}, which are crucial for accurate pose prediction, the features $h$ act as a bridge between these detailed representations and the higher-level measure of relative pose. Finally, similar to VLocNet++~\cite{radwan2018vlocnet++}, we concatenate
features $h^P_i$ and $h^P_j$, and pass them to a pose 
classification head
$C^P$, which computes their inter-feature correlation~\cite{cai2021extreme} and outputs a relative pose estimate $P_{i, j}$. 
\SM{Formally, $P_{i, j} = C^P([h^P_i, h^P_j]) $.}
The estimates are used in an auxiliary loss defined below.
See Fig.~\ref{fig:model} bottom-right.

\subsection{Model training}\label{sec:training}
\paragraph{Video captioner training.} 
We train the off-the-shelf video captioners~\citep{zhang2023video, li2023mvbench} in the best view pseudo-labeler $L$ (cf.~Sec.~\ref{sec:labeler})
by initializing them with the pretrained parameters released by their authors, and subsequently finetuning them on our multi-view videos
with the standard negative log-likelihood loss~\cite{zhang2023video, li2023mvbench}. This helps us leverage the knowledge from internet-scale pretraining and improve the captioning performance.

\paragraph{View selector training.} 
We train our view selector 
with a combination of two losses: a) a view 
classification
loss $\mathcal{L}^W$, and b) a relative camera pose prediction loss $\mathcal{L}^P$. While $\mathcal{L}^W$ provides a direct learning signal to the view 
classifier
$W$, $\mathcal{L}^P$ helps its visual encoder 
$F$ generate visual features that capture the important
inter-viewpoint
differences.

We propose a novel loss formulation for $\mathcal{L}^W$, which accounts for the cases where there is more than one best view pseudo-label in 
$\mathcal{B}$ (cf.~Sec.\ref{sec:labeler}). 
Specifically, we set $\mathcal{L}^W$ to 
\begin{equation}
    \mathcal{L}^W = min\{\mathcal{L}_{CE}(\tilde{\mathcal{B}}, \hat{\mathcal{B}}) \quad \forall \quad \hat{\mathcal{B}} \in \mathcal{B} \},
\end{equation}
where 
$\tilde{\mathcal{B}}$
is our best view estimate (cf.~Sec.~\ref{sec:selector})
and $\mathcal{L}_{CE}$ denotes the cross-entropy loss. Our formulation for $\mathcal{L}^W$ 
encourages our view 
classifier to
learn to predict as its estimate whichever among the 
pseudo-labels it
finds the easiest to predict~\cite{neitz2018adaptive, jayaraman2018time, gregor2018temporal}, 
thereby stabilizing training and improving task performance, as we show in results.

We set the relative camera pose prediction loss $\mathcal{L}^P$ to 
\begin{equation}
    \mathcal{L}^P = \frac{1}{N^2} {\sum_{(i, j) \in  \mathbb{N}}{\mathcal{L}_{CE}^P(\mathcal{P}_{(i, j)}, \mathcal{P}^{*}_{(i, j)})}}.
\end{equation}
Here, $\mathcal{L}^P_{CE}$ is the average cross-entropy loss over all discretized angles in 
$\mathcal{P}^*$ (cf. Sec.~\ref{sec:selector}), and $\mathbb{N} = \{1, \ldots, N\} \times \{1, \ldots, N\}$ is the set of all possible view pairs.  

We set our final training loss $\mathcal{L}^S$ for the view selector to $ \mathcal{L}^S = \mathcal{L}^W + w * \mathcal{L}^P$, where $w$ is the weight on the pose prediction loss,
\SM{and jointly train $W$ and $P$}.

\section{Experiments}\label{sec:experiments}

We overview all setup details and then provide results.

\subsection{Experimental setup}\label{sec:exp_setup}

\paragraph{Dataset.} 
We evaluate 
our model 
on two 
multi-view instructional video
datasets: 
Ego-Exo4D~\cite{grauman2023ego} and LEMMA~\cite{jia2020LEMMA}\footnote{Note that other instructional video datasets like  
HowTo100M~\cite{miech2019howto100m}, 
 CrossTask~\cite{zhukov2019cross}, and COIN~\cite{tang2019coin} are single-view and thus not suitable.}.  Ego-Exo4D contains
both physical (\eg basketball, dancing) and procedural (\eg cooking, bike repair) activities
with each video containing 5 time-synced
views---one is 
egocentric (ego)
and recorded by the participant with a 
headworn
camera, and the remaining views are 
exocentric (exo)
and recorded with stationary cameras 
kept
around the scene. 
LEMMA has videos of people performing 
household activities (\eg making juice, watering plant), where each video has 
the participant's ego view
and an exo view from a fixed camera placed facing the activity. 
Both have narrations: the Ego-Exo4D 
annotators 
provide temporally dense written descriptions of the participants' actions and relevant objects involved in the activity, while the LEMMA annotators
describe action verbs and objects being interacted with using a pre-defined vocabulary.
These two datasets let us evaluate different 
exo view
setups 
(multiple in EgoExo-4D vs.~single in LEMMA)
and diverse activity scenarios 
(physical and procedural in Ego-Exo4D vs.~household in LEMMA).
Ego-Exo4D and LEMMA provide a total of 86  and 20 hours of video data,
\SM{resulting in 648,665 and 63,538 clip-narration pairs,} respectively. 
See Supp.~for details,  including our clip segmentation strategy.

\paragraph{Implementation.} 
We use $K=3$ captioners in our 
pseudo-labeler: Video-Llama~\cite{zhang2023video} with Llama-2-Chat~\cite{touvron2023llama2} or Vicuna~\cite{vicuna2023} as the LLM decoder, and VideoChat2~\cite{li2023mvbench}, respectively, and the CIDEr~\cite{Vedantam2014CIDErCI} captioning metric to score views. 
\SM{We finetune the captioners on our datasets before using them to score views.} 
On the basis of a disjoint validation set, we set the bin size to 
$30$ degrees 
for generating relative camera pose labels (cf.~Sec.~\ref{sec:selector}), and the weight on the pose prediction loss to $w = 0.5$. 
\SM{See Supp. for 
more
details.}

\paragraph{Baselines.}

We compare against the following baselines and state-of-the-art methods:
\begin{itemize}[leftmargin=*]
\itemsep0em 
    \item \textbf{Ego-only, Random, Random-exo:} a set of naive baselines that predict the ego view (\textbf{Ego-only}), or a view randomly chosen from just exo (\textbf{Random-exo}) or all (\textbf{Random}) views, as the best view.
    \item \textbf{Hand-object, Body-area, Joint-count:} 
    a set of baselines that predict the view with the highest hand and object detection confidence per a state-of-the-art hand and object detector~\cite{cheng2023towards} (\textbf{Hand-object}), or the largest body area (\textbf{Body-area}) or joint count (\textbf{Joint-count}) per a state-of-the-art body pose detector~\cite{Jiang2023RTMPoseRM}, as the best view.
    \item \textbf{Snap angles
    ~\cite{Xiong_2018_ECCV, 10.1145/3183794}:} an automatic cinematography method that predicts the view with the highest foreground pixel count as the best view. We upgrade it to use today's SOTA segmentation models~\cite{liu2023grounding, kirillov2023segment}.
    \item  \textbf{Longest-caption:} 
    \KG{a baseline that \SM{uses our finetuned Video-Llama captioner~\cite{zhang2023video} to}
    \SM{predict}
    captions for each input view, and selects the one that produces the longest narration as the best view.
    Intuitively, here caption length is used a proxy for informativeness.  Recall that our model does not infer captions on test data.}

\end{itemize}

While the first set of baselines are naive heuristics to test if intelligent view selection is even necessary, the second set  accounts for the prior that visibility of people and person-object interactions are strongly linked to the informativeness of a view, and the third represents the most relevant existing methods in the literature. 
\KG{Finally, the Longest-caption baseline offers an alternative, more naive way to incorporate language for our task.}

\paragraph{Evaluation metrics.}

We perform both \emph{automatic} and \emph{human} evaluation. 
For automatic evaluation, we 
measure how well the selected view predicts two things: narrations and action/object terms.
For the former, we use a state-of-the-art video captioner~\cite{zhang2023video, li2023mvbench} to predict the narrations given our chosen views, and then compare the predictions with the view-agnostic ground-truth narrations through standard captioning metrics: 
\textbf{\cider}~\cite{Vedantam2014CIDErCI} and \textbf{\meteor}~\cite{banerjee-lavie-2005-meteor}.
For the latter,
we use 1) \textbf{Verb \iou~(V-\iou)} , 2) \textbf{Noun \iou~(N-\iou)}, and 3) \textbf{Noun-chunk \iou~(NC-\iou)}, which measure the overlap in the sets of verbs, nouns, and noun chunks between the 
ground-truth 
and predicted
narrations, 
where noun chunks are nouns grouped with their modifiers (\eg adjective, article).
In short, the more the view  deemed as ``best" by our model predicts things consistent with the comprehensive view-agnostic ground truth, the better it is. 
\KG{We stress that this suite of metrics, together with the human evaluation below, goes beyond CIDEr (used in our language-guided training) to gauge the quality of the selected views.}

Through \textbf{human evaluation}, we 
assess two important
model
aspects:
1)
pseudo-label
quality,
and 2)
view selection performance.
We conduct both assessments 
through 
pairwise
comparison
of views, which reduces the cognitive load
on human judges and increases their reliability~\cite{10.1371/journal.pone.0190393, agrawal-etal-2023-findings, goncalves2024peavs}.
Given a view pair, the human judge
can select either 
view
or both, depending on if they prefer one over the
other specifically for the purpose of activity understanding (i.e., a \emph{win} for the preferred view,
and a \emph{loss} for the other), 
or find them equally 
informative (\emph{tie}). 
Critically,
we do not 
show
the evaluators 
the ground-truth narrations.
In other words, our study \emph{directly evaluates the human-preferred 
viewpoints}---independent of narrations---and hence is unbiased by the fact our model leverages language during training.
We obtain human judgments on 1) pseudo-label quality by 
pairing
the best and the worst views per our pseudo-labeler (cf.~Sec.~\ref{sec:labeler}),
and 
2) view prediction quality by 
pairing the views predicted by our 
and 
the
best
baseline's predicted views. 
We decide the view order in each pair randomly.
We 
do
each study for both datasets with 10 participants and 
70 randomly chosen 
test
view pairs. Our inter-evaluator agreement rate is 78.5\%.

\subsection{Results}\label{sec:results}
\paragraph{Automatic evaluation.}

\begin{table*}[!t]
\small
  \centering
  \setlength{\tabcolsep}{4pt}
  \resizebox{0.9\linewidth}{!}{
    \begin{tabular}{l c c c c c| c c c c c}
    \toprule
     &  \multicolumn{5}{ c| }{Ego-Exo4D~\citep{grauman2023ego}} & \multicolumn{5}{ c }{LEMMA~\citep{jia2020LEMMA}} \\
     &  \multicolumn{2}{ c }{\textit{Captioning}} & \multicolumn{3}{ c| }{\textit{Actions and objects}} & \multicolumn{2}{ c }{\textit{Captioning}} & \multicolumn{3}{ c}{\textit{Actions and objects}}\\
     Model & \cider~\cite{Vedantam2014CIDErCI} & \meteor~\cite{banerjee-lavie-2005-meteor} & V-\iou & N-\iou & NC-\iou & \cider~\cite{Vedantam2014CIDErCI} & \meteor~\cite{banerjee-lavie-2005-meteor} & V-\iou & N-\iou & NC-\iou\\
    \midrule
    Ego-only & 12.2 & 47.2 & 32.2 & 36.7 & 30.6 & 41.7 & 71.1 & 38.2 & 41.3 & 17.5\\
    Random & 11.5 & 45.9 & 30.4 & 36.6 & 31.0 & 30.9 & 63.1 & 31.2 & 33.2 & 12.8\\
    Random-exo & 11.9 & 46.0 & 30.5 & 37.0 & 30.9 & 17.7 & 51.3 & 21.6 & 22.4 & 6.8\\
    Hand-object & 12.6 & 47.4 & 33.6 & 36.7 & 29.6 & 40.7 & 72.7 & 38.5 & 41.5 & 17.9\\
    Body-area & 12.9 & 48.2 & 32.5 & 37.2 & 31.1 & 42.1 & 73.8 & 38.6 & 41.3 & 17.6\\
    Joint-count & 12.6 & 46.6 & 31.5 & 29.1 & 27.7 & 17.8 & 51.4 & 21.7 & 22.4 & 6.7\\
    Snap angles~\cite{Xiong_2018_ECCV, 10.1145/3183794} & 12.2 & 46.7 & 30.7 & 35.8 & 29.1 & 38.9 & 70.6 & 37.1 & 40.2 & 17.1\\
    Longest-caption & 10.7 & 47.3 & 30.5 & 34.6 & 28.8 & 32.7 & 65.4 & 36.9 & 37.9 & 15.3 \\
    \textbf{\textsc{LangView} (Ours)} & \textbf{13.5} & \textbf{48.4} & \textbf{33.7} & \textbf{39.2} & \textbf{32.9} & \textbf{42.7} & \textbf{74.4} & \textbf{40.1} & \textbf{42.9} & \textbf{18.9}\\
    \bottomrule
  \end{tabular}
  }
  \vspace{-0.2cm}
  \caption{
  View selection results. 
  All metrics are in $\%$ and higher is better.
  }
  \label{tab:auto_eval_egExAll}
  \vspace*{-0.5cm}
\end{table*}

Table~\ref{tab:auto_eval_egExAll} top shows our results for automatic evaluation. The naive baselines are generally the worst performers, indicating that blindly choosing the ego view at all times, or picking a random viewpoint is not enough for our challenging task. Interestingly, while random-exo improves over ego-only and random on 
Ego-Exo4D~\cite{grauman2023ego}, it fares worse on LEMMA~\cite{jia2020LEMMA}, possibly because 
activities like rock climbing or
basketball in Ego-Exo4D 
involve more head and body motion than the household activities (\eg cooking, watching TV)
in LEMMA, and consequently, require stationary 
exo
cameras for better coverage. Using 
intelligent 
heuristics like hand and object detection confidence (Hand-object), 
body visibility (Joint-count and Body-area), or foreground 
object prevalence (Snap angles~\cite{Xiong_2018_ECCV, 10.1145/3183794}) generally improve task performance, showing that 
these 
methods provide useful cues for our task.
\SM{However, the Longest-caption baseline generally underperforms the naive baselines, possibly because the view with the longest predicted narration provides excessive details about the scene, which are irrelevant to the activity. 
}

Our model significantly outperforms all baselines across metrics, on both datasets. It shows our idea of 
pseudo-labeling
by leveraging 
the
view-dependent 
quality
of predicted narrations is effective in practice. 
Furthermore, our model's improvement over the heuristics illustrates that 
our language-guided training facilitates
complex but essential reasoning about the interplay between human actors and interacting objects, more than what is possible with hand and object, or body pose detectors. Finally, our superior performance on both datasets underlines the efficacy of our design and its ability to generalize to different
activity types---both physical and non-physical in Ego-Exo4D vs. 
household
in LEMMA---
and camera setups---single 
vs. multiple exo cameras in LEMMA and Ego-Exo4D, respectively. See Supp. for 
results on 
the single exo camera 
and 3-fold evaluation with Ego-Exo4D.

\paragraph{Human evaluation.}

\begin{table}[!t]
\small
  \centering
  \setlength{\tabcolsep}{3pt}
    \begin{tabular}{l c c c}
    \toprule

    Assessment type &  \multicolumn{1}{ c }{Win} & \multicolumn{1}{ c}{Loss} & \multicolumn{1}{ c }{Tie}\\
    \midrule
    \multicolumn{1}{ l }{\textit{Pseudo-label: best vs. worst view}}\\
    \quad Ego-Exo4D~\cite{grauman2023ego} & \textbf{53.3} & 28.9 & 17.8\\
     \quad LEMMA~\cite{jia2020LEMMA} & \textbf{46.7} & 38.9 & 14.4\\
    \multicolumn{1}{ l }{\textit{View prediction: ours vs best baselines}}\\
    \quad \textbf{Ego-Exo4D~\cite{grauman2023ego}} \\
    \quad\quad Ours vs. 
    Hand-object & \textbf{52.2} & 40.0 & 7.8\\
    \quad\quad Ours vs. 
    Body-area  & \textbf{55.1} & 39.3 & 5.6\\
    \quad \textbf{LEMMA~\cite{jia2020LEMMA}} \\
    \quad\quad Ours vs. 
    Hand-object & \textbf{43.3} & 41.1 & 15.6\\
    \quad\quad Ours vs. 
    Body-area & \textbf{46.7} & 35.5 & 17.8\\
    \bottomrule
  \end{tabular}
  \vspace{-0.2cm}
  \caption{Human evaluation results for our 
  pseudo-label and view prediction quality. 
  All values are in $\%$. Significance, $p \leq 0.05$.}
  \label{tab:human_eval}
  \vspace*{-0.5cm}
\end{table}

Table~\ref{tab:human_eval} shows our human evaluation results using \textit{win}, \textit{loss} and \textit{tie} percentages.
In our pseudo-label quality study, the human preference for the best view per our pseudo-labels is significantly higher than the worst view. Given that the participants were asked to base their responses on the views' suitability for 
activity understanding, this validates our hypothesis that the view-specific 
quality 
of predicted narrations
is correlated with human 
preference 
for informative views,
and can be exploited for 
sourcing 
best view pseudo-labels.
Furthermore, a significant \emph{tie} rate indicates a considerable presence of instances with multiple high-quality views, re-emphasizing the challenging nature of our task.

In our view 
selection
assessment, we find  that our 
selected views are
preferred significantly more
than the 
two 
top
baselines on both datasets.
We achieve a strong win rate 
boost of at least $\sim$11\% over the baselines across different evaluation scenarios.
The only exception is our
model vs.
Hand-object~\cite{cheng2023towards} on LEMMA, where the improvement margin 
is a more modest $\sim$2\%, likely because the view that best shows 
hand-object interactions is a reasonably good view for showing LEMMA-style activity (cooking, watering plant, etc.).
These results
show
that our 
weakly-supervised method
is 
better capable of implicitly modeling human view preference and using this model to perform 
more accurate
view selection. Finally, our consistent performance gains on human evaluation reinforce  
our
automatic evaluation
metrics, and all model analysis we do using them.


\paragraph{Qualitative examples.}
\begin{figure*}[!tb] 

    \centering
    \includegraphics[width=0.9\linewidth]{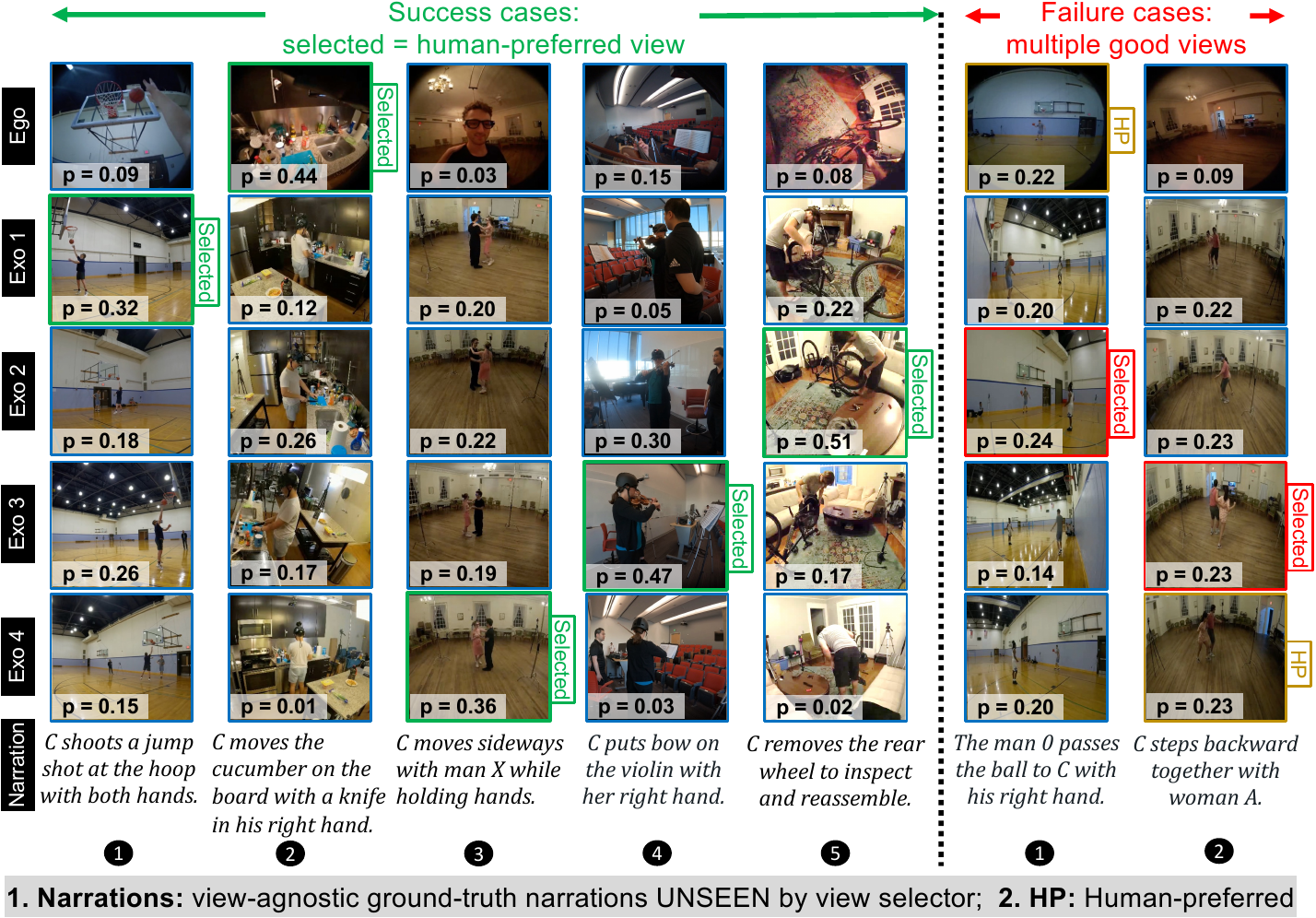}
    \caption{\textbf{Left:} sample successful predictions 
    by
    our view selector. 
    For each clip, our model chooses the view that shows the action, and the objects and body parts involved in it, 
    most
    clearly\SM{,} 
    and hence, is 
    most
    informative.
    \textbf{Right}: Sample failure cases for our model, where there are multiple high-quality views that differ only in certain nuances, which are discernible by a human but not our model trained through narration guidance. Whereas humans prefer a view that better captures the direction of the ball towards the camera-wearer in sample 1, or shows the full backward motion of the dancers in sample 2, our model 
    choose a view that 
    shows all entities
    mentioned in the narration.
    } 

\label{fig:qual}
\vspace{-0.35cm}
\end{figure*}

Fig.~\ref{fig:qual} (left) shows
some
success cases.
Note how our model chooses 
views that clearly show all the essential elements of the actions, including
the different entities involved and
and  
their motion.
E.g., 
in clip 2, our model chooses the ego view as it clearly shows the horizontal 
knife 
motion
and vegetables on the 
cutting
board, whereas in clip 3, it selects an exo view that best shows the joined hands and the sideways movement of the dancers.
Thus, depending on 
the activity,
our model adaptively chooses a view that accurately 
shows
its important details. See Supp.~video for more examples.

We also observe two common failure types. 
The 
first type occurs when  
\SM{there are} multiple high-quality views. 
See Fig.~\ref{fig:qual} (right).
In the 
second one,
our model 
chooses different 
views
for 
very similar activities, which involve very similar types, positions and motion of relevant objects and body parts, even when the 
views
are not 
equally good. 
This occurs possibly because our selector occasionally picks on spurious cues that are not as viewpoint-dependent as the activity itself but also do not help with its understanding.

\begin{table}[!t]
\small
  \centering
  \setlength{\tabcolsep}{2pt}
    \begin{tabular}{l c c c c c}
    \toprule
     &  \multicolumn{2}{ c }{\textit{Captioning}} & \multicolumn{3}{ c }{\textit{Actions and objects}}\\
     Model & \cider & \meteor & V-\iou & N-\iou & NC-\iou \\
    \midrule
    Ours w/o $P$ & 13.2 & \textbf{49.2} & 33.4 & 38.7 & 32.8 \\
    Ours w/o $\mathcal{L}^W$ & 13.5 & 48.8 & 34.0 & 37.4 & 32.3 \\
    Ours w/o $G$ & \textbf{13.7} & 48.7  & \textbf{34.3} & 38.5 & 32.8\\
    Ours w/o $G$, $\mathcal{L}^W$, $P$ & 13.2 & 48.0 & 33.5 & 37.7 & 32.5 \\
    \textbf{Ours} & 13.5 & 48.4 & 33.7 & \textbf{39.2} & \textbf{32.9}\\

    \bottomrule
  \end{tabular}
  \vspace{-0.2cm}
  \caption{
  Ablation results on the large-scale Ego-Exo4D~\cite{grauman2023ego} dataset.
  $G$ denotes the 
  rank aggregator
  in our 
  pseudo-labeler. 
  All metrics are in $\%$. Significance, $p \leq 0.05$.
  }
  \label{tab:ablations_autoMetrics_Ego-Exo4D}
  \vspace*{-0.5cm}
\end{table}

\paragraph{Ablations.}
In 
Table~\ref{tab:ablations_autoMetrics_Ego-Exo4D} (top)
we report our model ablation results on the large-scale 
Ego-Exo4D~\cite{grauman2023ego} dataset. 
Not predicting inter-view camera pose in training
significantly hurts 
performance on all metrics except \meteor~\cite{banerjee-lavie-2005-meteor}.
This shows that our pose predictor enhances the selector's view sensitivity.
Training our model
with a standard cross-entropy loss 
by
randomly 
choosing
a sample
from our 
pseudo-label set, 
instead of our proposed loss (c.f.~Sec.~\ref{sec:training})
hurts performance on N- and NC-IoU. 
This occurs possibly because the captioners in our pseudo-labeler 
sometimes hallucinate 
and add 
less informative
views to the pseudo-label set,
which when sampled causes 
training instability.
Removing the 
rank aggregator to obtain pseudo-label consensus 
also negatively impacts 
performance on N- and NC-IoU, 
as having multiple captioners vote on the best view reduces captioning noise
and improves pseudo-label quality. Finally, removing all three components 
consistently
degrades 
performance
across
metrics, showing that the model needs \emph{at least one 
component}
to 
perform well
on different metrics.

See Supp. for additional ablations, and  analyses of the view dependence of our visual features, 
our pseudo-labeler,
and the impact of the rank of our selector's sampled view on view selection performance, and our model's attention maps.

\section{Conclusion}
We tackle view selection in multi-view instructional videos in the absence of best view labels. To that end, we design a novel framework composed of a best view pseudo-labeler that uses the view-dependent 
quality of estimates of
video descriptions to automatically generate best view pseudo-labels, and a best view selector that given a video, produces a 
best view prediction.
Our method significantly outperforms several state-of-the-art baselines on two challenging multi-view instructional video datasets. In future work, we will explore future best-view anticipation for improving the energy efficiency of multi-view instructional video capture setups.
\small{\noindent \textbf{Acknowledgements:} UT Austin is supported in part by the IFML NSF AI Institute. KG is paid as a research scientist by Meta, and SM was a visiting researcher at the same when this work was done.}

{
    \small
    \bibliographystyle{ieeenat_fullname}
    \bibliography{mybib}
}

\clearpage
\section{Supplementary material}
In this supplementary material we provide additional details
about:
\begin{itemize}
    \item Video (with audio) for qualitative illustration of our task and
    qualitative assessment of our view predictions (Sec.~\ref{sec:supp_vid}), as referenced in `Qualitative examples' in Sec.~\ref{sec:results} in main
    \item Additional ablations of our model components (Sec.~\ref{sec:ablations_supp}), as mentioned in `Ablations' in Sec.~\ref{sec:results} in main
    \item Analysis of the view-specificity of our model's learned visual features (Sec.~\ref{sec:supp_tsne}), as noted in `Ablations' in Sec.~\ref{sec:results} in main
    \item Analysis of the impact of rank our selector's sampled view on view selection performance (Sec.~\ref{sec:supp_sampledRankEffect}), as mentioned in `Ablations' in Sec.~\ref{sec:results} in main
    \item Examples of our view selector's attention heatmaps (Sec.~\ref{sec:supp_attnHeatMaps}), as noted in `Ablations' in Sec.~\ref{sec:results} in main
    \item Analysis of our pseudo-labeler (Sec.~\ref{sec:supp+pseudolabelerAnalysis}), as referenced in Sec.~\ref{sec:results} in main
    \item View selection results on Ego-Exo4D~\cite{grauman2023ego} with a single exo camera (Sec.~\ref{sec:supp_egoExo_singleExo}), as mentioned in Sec.~\ref{sec:results} in main
    \item 3-fold evaluation of our view selector on Ego-Exo4D~\cite{grauman2023ego}, as noted in `Automatic evaluation' in Sec.~\ref{sec:results} in main

    \item Analysis of the relation between our model performance and the distribution of different concepts in the ground-truth train narrations (Sec.~\ref{sec:supp_perfVsNarrationConceptDist})
    
    \item Our pseudo-labeling cost (Sec.~\ref{sec:supp_pseudoLabelingCost})
    \item Dataset details (Sec.~\ref{sec:supp_dataset}) in addition to what is provided in Sec.~\ref{sec:exp_setup} in main
    \item 
    Implementation details (Sec.~\ref{sec:supp_implementation}), as noted in Sec.~\ref{sec:exp_setup} in  main
\end{itemize}

\subsection{Supplementary video}\label{sec:supp_vid}
The supplementary video qualitatively depicts our task of \SM{view-selection in multi-view instructional videos}.
Moreover, we qualitatively \SM{illustrate our key idea}, \SM{Language for Weakly Supervising View Selection}, show our model’s view selection quality at the level of both individual clips and long videos (comprising multiple clips), and compare our predictions with those of two best-performing baselines. Some long videos also have the audio commentary of the participant. Please use headphones
to hear the audio correctly. 
The video is available on \url{http://vision.cs.utexas.edu/projects/which-view-shows-it-best}.

\subsection{Additional ablations}\label{sec:ablations_supp}
\begin{table*}[!t]
\small
  \centering
  \setlength{\tabcolsep}{2pt}
  \resizebox{\linewidth}{!}{
    \begin{tabular}{l c c c c c}
    \toprule
     &  \multicolumn{2}{ c }{\textit{Captioning}} & \multicolumn{3}{ c }{\textit{Actions and objects}}\\
     Model & \cider~\cite{Vedantam2014CIDErCI} & \meteor~\cite{banerjee-lavie-2005-meteor} & V-\iou & N-\iou & NC-\iou \\
    \midrule
    Ours w/o captioner finetuning in our pseudo-labeler $L$ & 0.4 & 12.2 & 1.4 & 6.5 & 4.8 \\
    Ours w/o direction prediction between camera centers in our relative camera pose predictor $P$ & 12.9 & 48.1 & 32.5 & 36.8 & 31.6 \\
    \textbf{Ours} & \textbf{13.5} & \textbf{48.4} & \textbf{33.7} & \textbf{39.2} & \textbf{32.9}\\
    \bottomrule
  \end{tabular}
  }
  \vspace{-0.2cm}
  \caption{Ablation results on the large-scale Ego-Exo4D~\cite{grauman2023ego} dataset, in addition to what is provided in 
  `Ablations' in
  Sec.~\ref{sec:results} 
  in main. For the ablation that does not predict the direction between camera centers during relative pose prediction, we predict the exact differences in locations between camera centers instead. Significance, $p \leq 0.05$.}
  \label{tab:ablations_autoMetrics_supp_Ego-Exo4D}
\end{table*}

In 
`Ablations' in 
Sec.~\ref{sec:results}
of main, we ablate different model components to understand their contribution to our view selection performance. Here, we provide additional ablations to further analyze our model. Table~\ref{tab:ablations_autoMetrics_supp_Ego-Exo4D} shows the results. Upon keeping the off-the-shelf captioners~\cite{zhang2023video, li2023mvbench} frozen when generating our best view pseudo-labels using our pseudo-labeler $L$ 
(Sec.~\ref{sec:captions} in main), 
the performance declines drastically, indicating that the generic captions generated by frozen off-the-shelf captioners are not at all suitable for activity understanding in instructional videos. Upon predicting the exact displacement of one camera center relative to another, instead of the rough direction between them, when predicting the inter-view relative poses using our relative camera pose predictor $P$ 
(Sec.~\ref{sec:selector} in main), 
we against observe a significant drop in view selection performance. This happens possibly because predicting the exact difference in locations between two camera centers can be intractable in our setting, due to the unknown scale of objects and background.

\subsection{View dependence of visual features}\label{sec:supp_tsne}
\begin{figure*}[!tb] 
    \centering
    \begin{subfigure}[b]{0.31\linewidth}
    \centering
    \includegraphics[width=\linewidth]{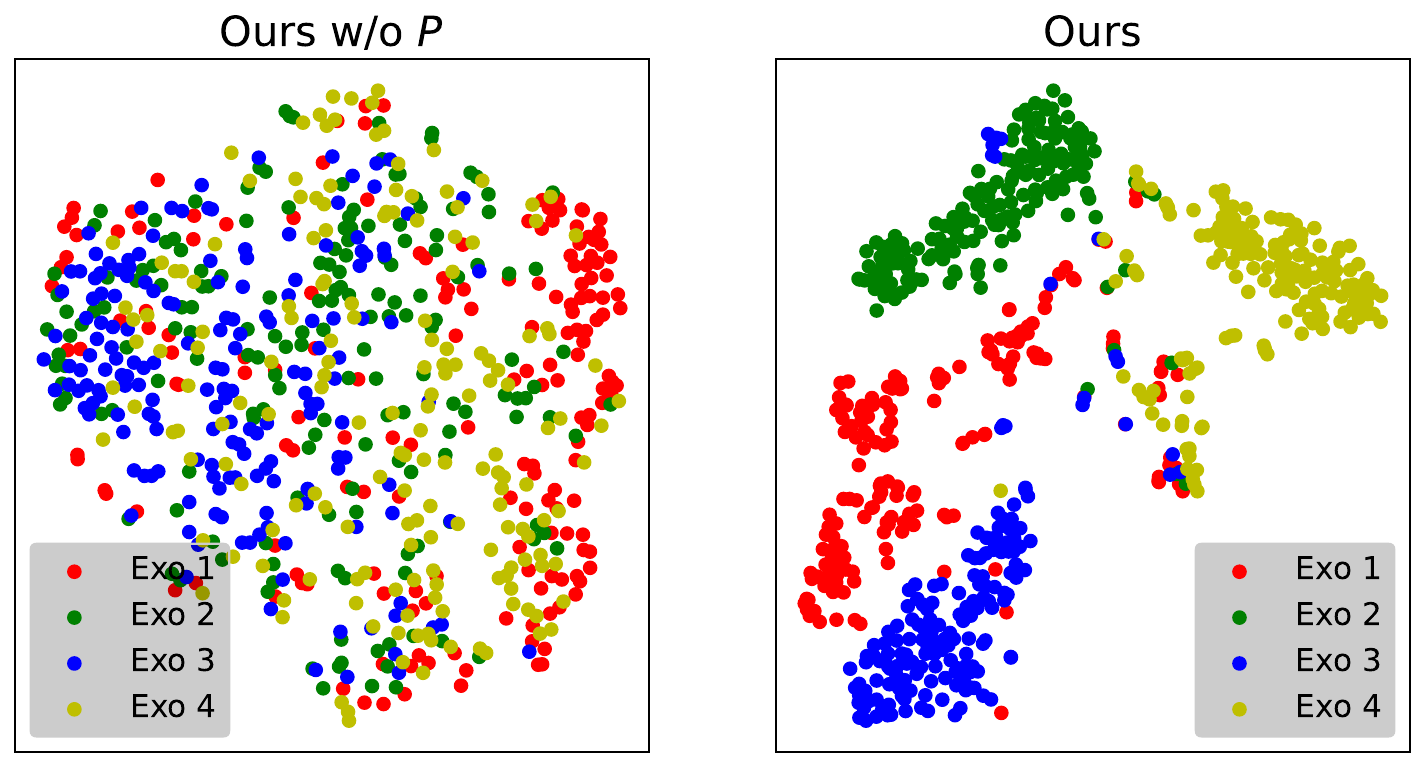}
    \caption{Basketball \#1}
    \label{fig:tsne_bb2}
    \end{subfigure}\hfill
    \begin{subfigure}[b]{0.31\linewidth}
    \centering
    \includegraphics[width=\linewidth]{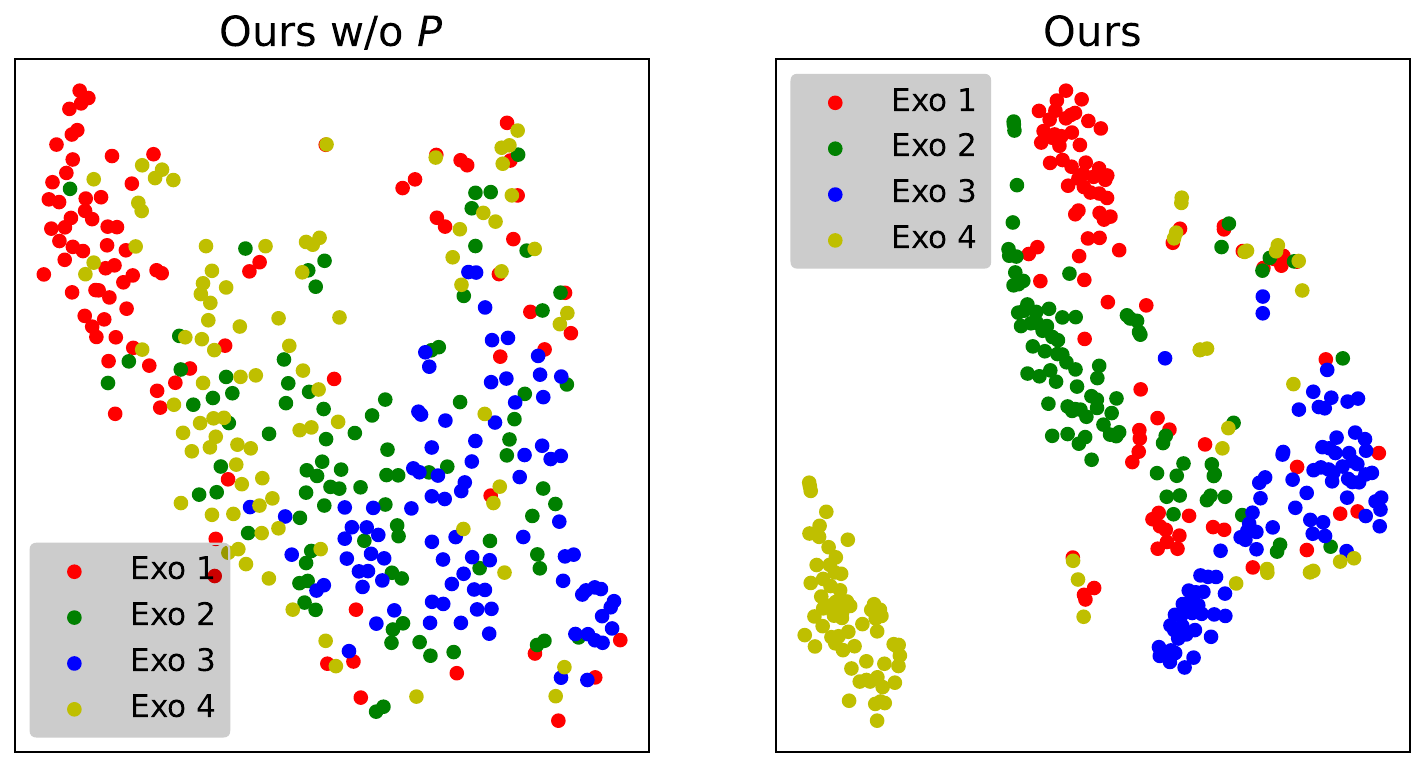}
    \caption{Basketball \#2}
    \label{fig:tsne_bb2}
    \end{subfigure}\hfill
    \begin{subfigure}[b]{0.31\linewidth}
    \centering
    \includegraphics[width=\linewidth]{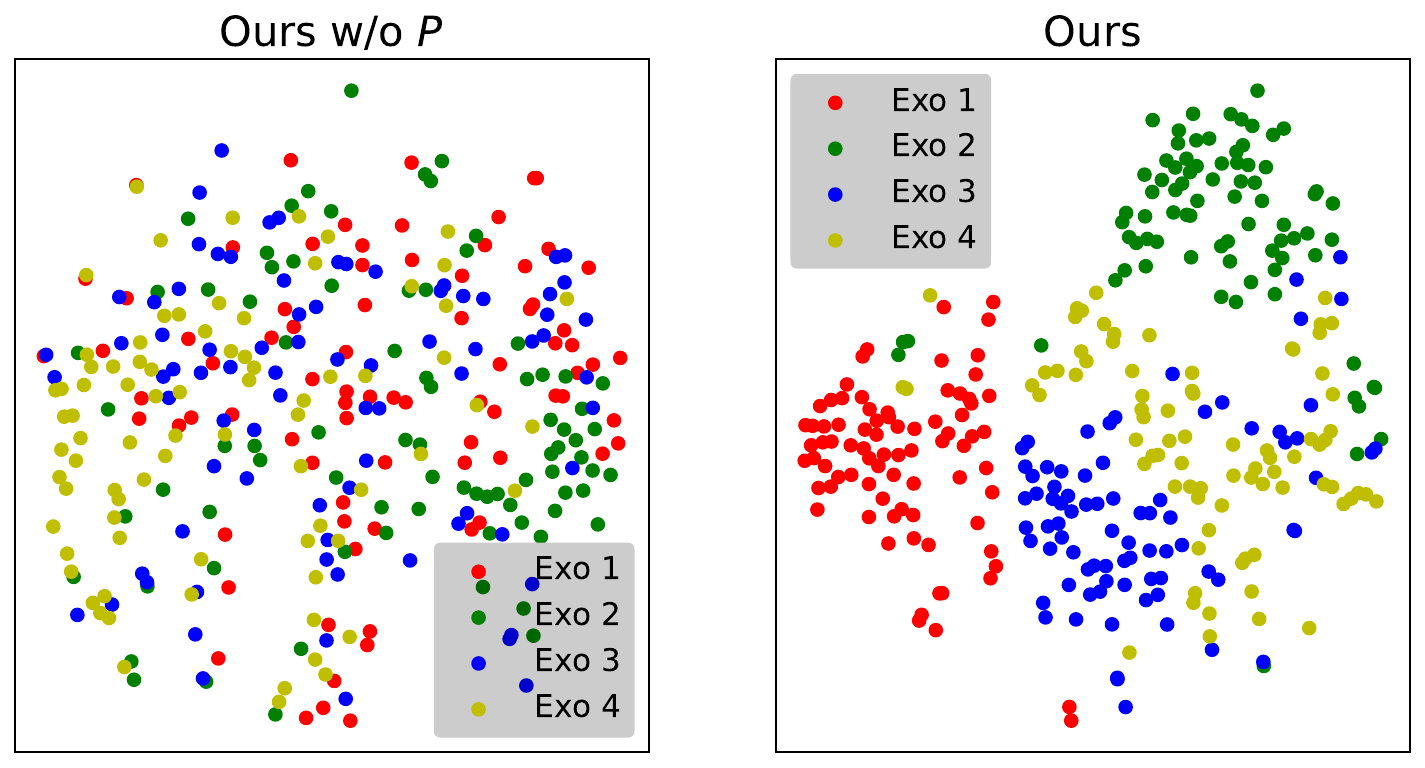}
    \caption{Dance \#1}
    \label{fig:tsne_dnc1}
    \end{subfigure}\hfill

    \centering    
    \begin{subfigure}[b]{0.31\linewidth}
    \centering
    \includegraphics[width=\linewidth]{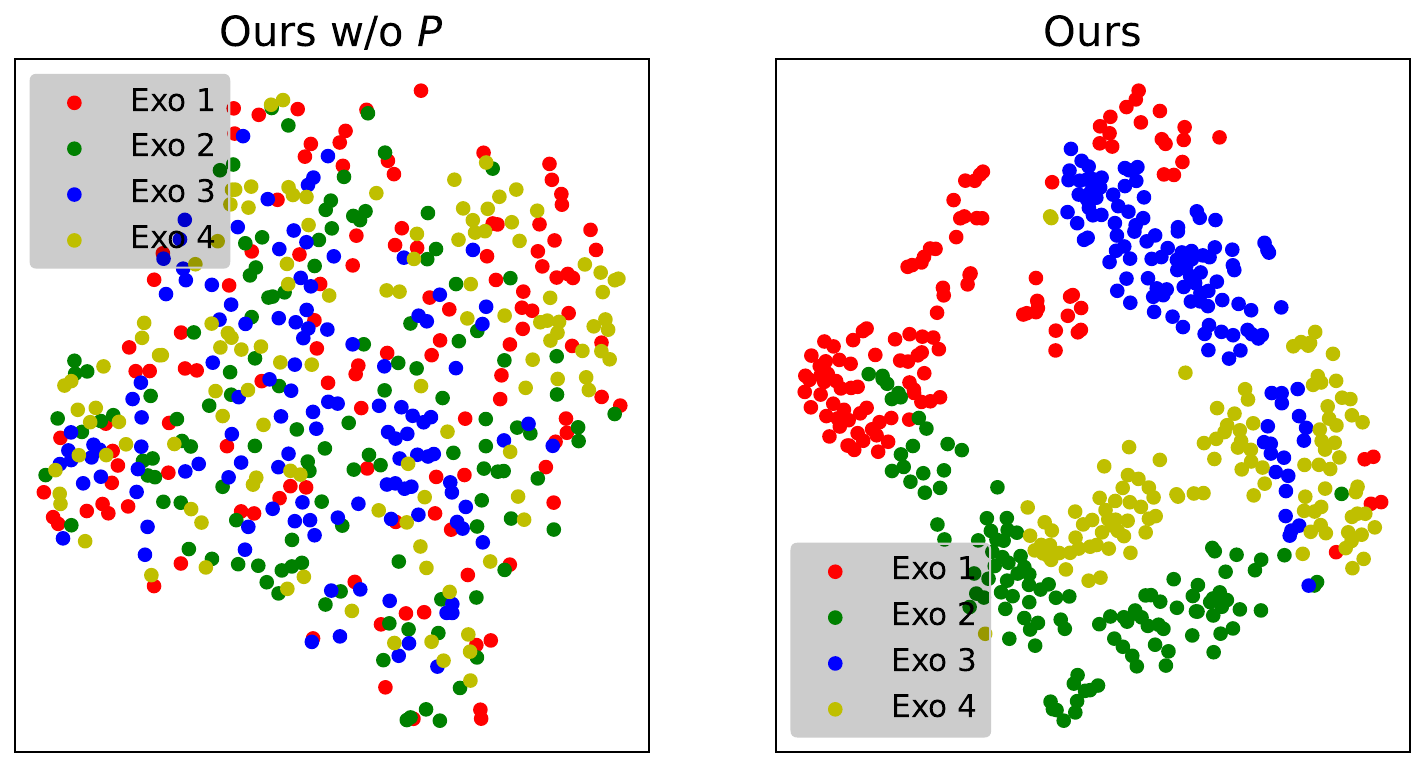}
    \caption{Dance \#2}
    \label{fig:tsne_dnc2}
    \end{subfigure}\hfill
    \begin{subfigure}[b]{0.31\linewidth}
    \centering
    \includegraphics[width=\linewidth]{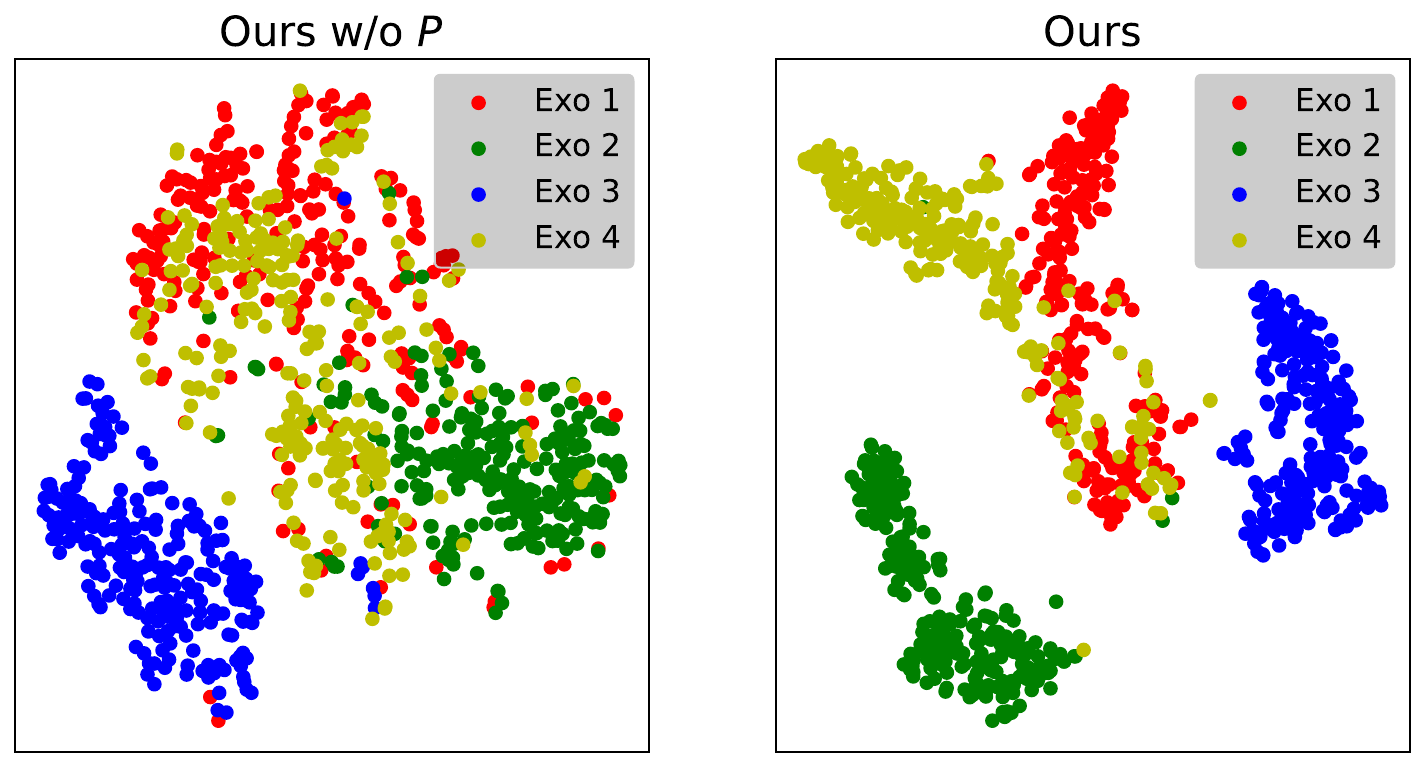}
    \caption{Bike repair \#1}
    \label{fig:tsne_bk1}
    \end{subfigure}\hfill
    \begin{subfigure}[b]{0.31\linewidth}
    \centering
    \includegraphics[width=\linewidth]{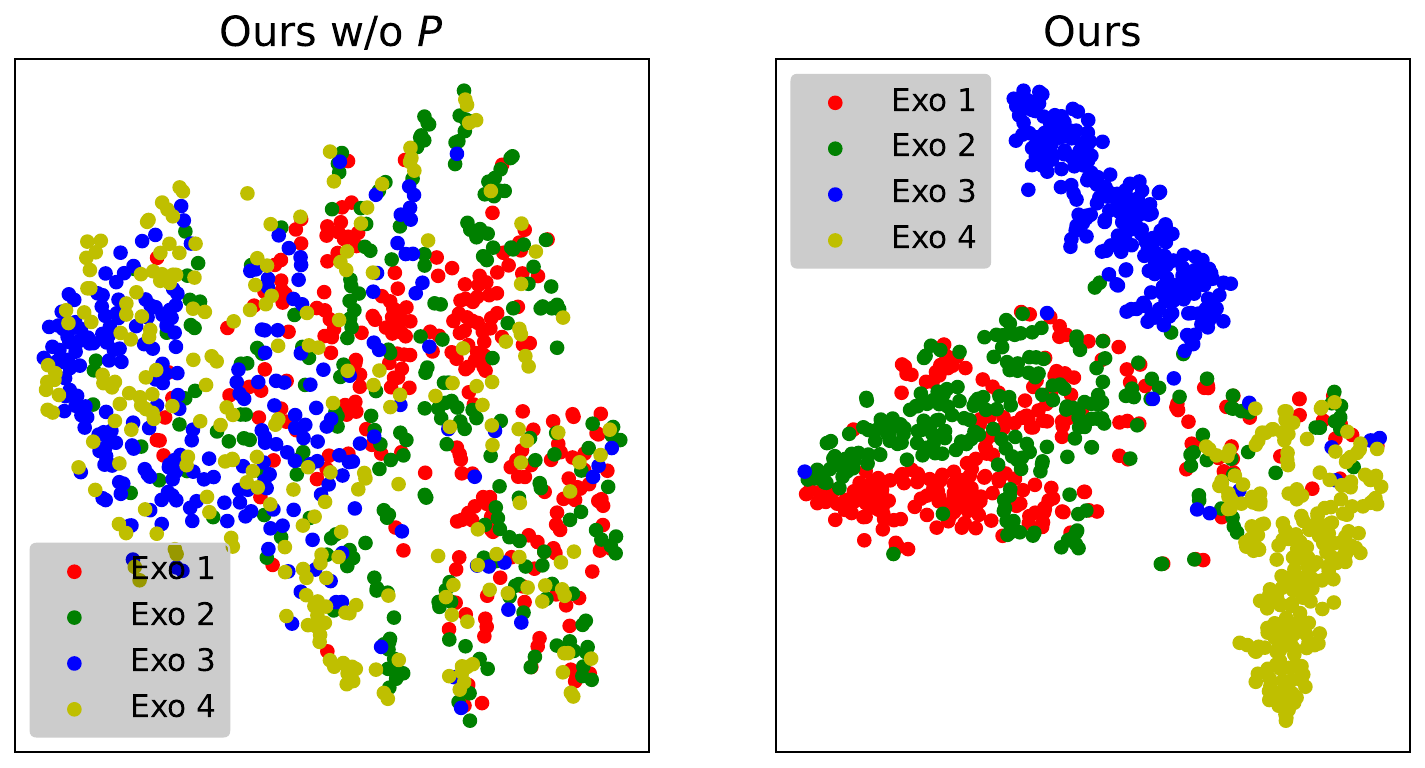}
    \caption{Bike repair \#2}
    \label{fig:tsne_bk2}
    \end{subfigure}\hfill

    \centering
    \begin{subfigure}[b]{0.31\linewidth}
    \centering
    \includegraphics[width=\linewidth]{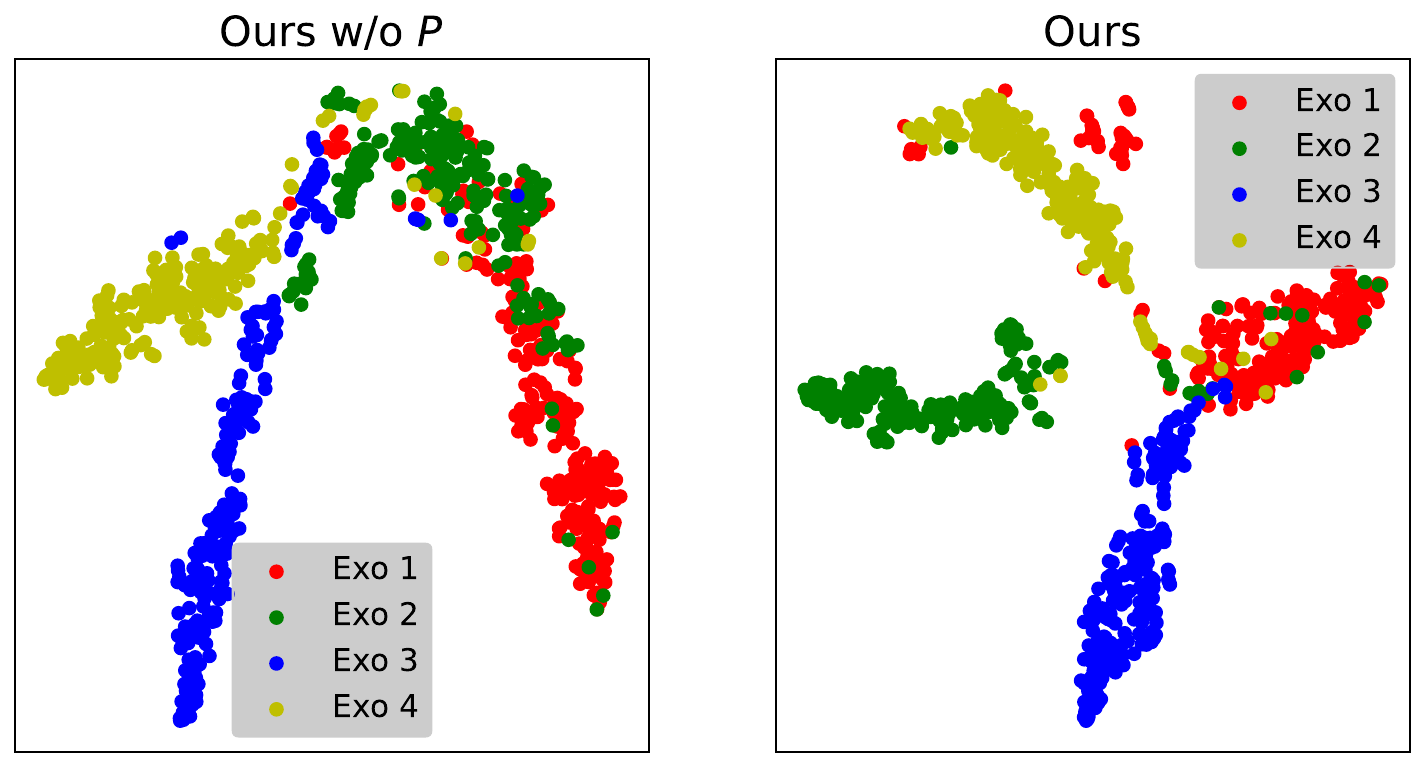}
    \caption{Cooking \#1}
    \label{fig:tsne_ckng1}
    \end{subfigure}
    \hspace{0.025\linewidth}
    \begin{subfigure}[b]{0.31\linewidth}
    \centering
    \includegraphics[width=\linewidth]{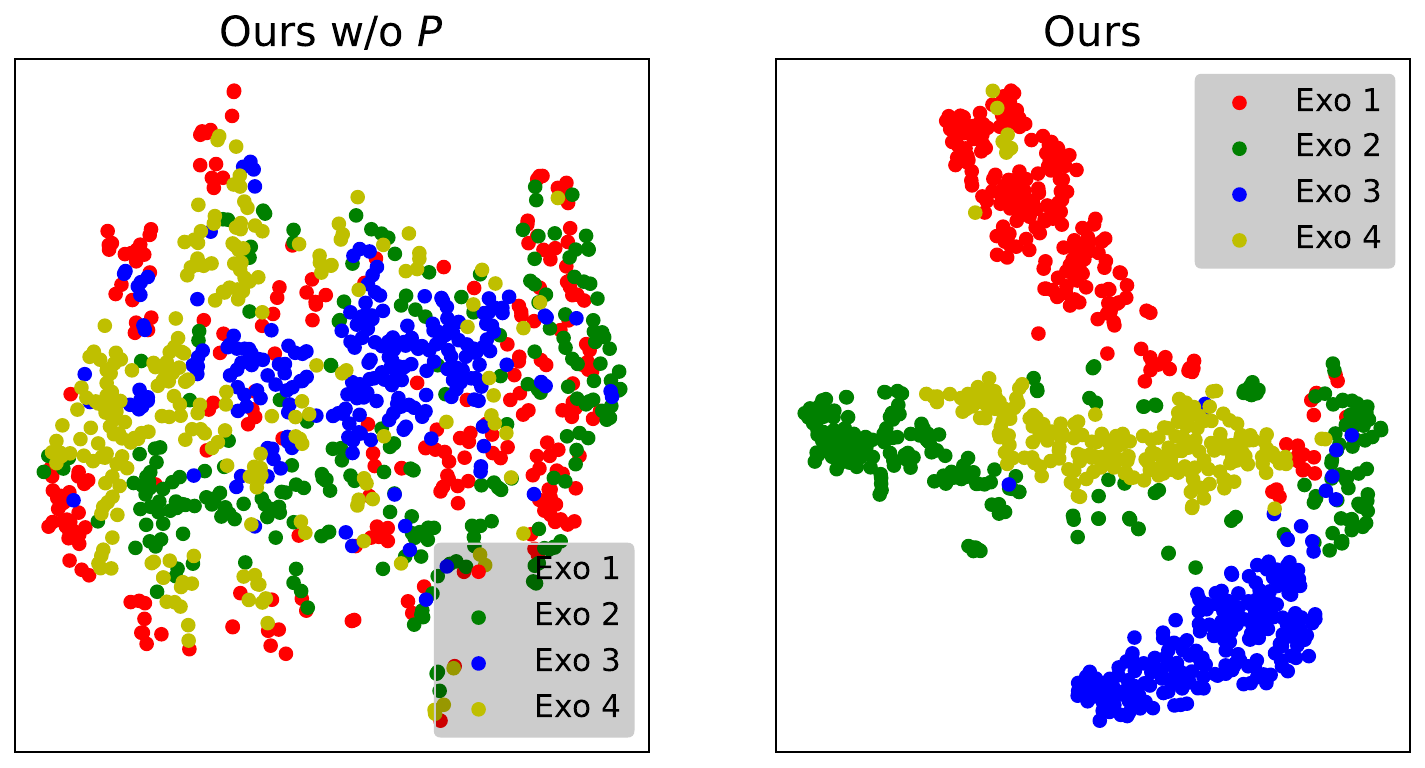}
    \caption{Cooking \#2}
    \label{fig:tsne_ckng2}
    \end{subfigure}\hfill

    \caption{
    t-SNE~\cite{vandermaaten08a} plots of 
    exo visual features of sample Ego-Exo4D~\cite{grauman2023ego} videos from basketball, bike repair, dance and cooking scenarios. Our model, when trained with the relative camera pose predictor, produces visual features that form neater clusters when grouped on the basis of different 
    exo views, highlighting their improved view sensitivity.}
\label{fig:supp_tsne}
\end{figure*}

Fig.~\ref{fig:supp_tsne}
shows the t-SNE visualizations of the visual features corresponding to the exo views of
videos from different scenarios--basketball, dance, bike repair and cooking.
The scenarios 
have varying levels of motion of the camera wearer's body and relevant objects--whereas basketball and dance involve moving large and fast movements of the full body and salient objects, bike repair and cooking primarily just involve hands and need less body and object motion. Our learned visual features for the exo cameras when grouped on the basis of the camera ID, produce tighter clusters across samples from different scenarios, compared to the model variant trained without our relative camera pose estimation loss (`View selector training' in Sec.~\ref{sec:selector} in main).
This demonstrates that our model's superior ability to learn view-dependent features cuts across different types of activity and different levels of body and object motion, which consequently leads to a stronger view selection performance.

\subsection{Sampled view rank}\label{sec:supp_sampledRankEffect}
\begin{table}[!t]
\small
  \centering
  \setlength{\tabcolsep}{2pt}
  \resizebox{\linewidth}{!}{
    \begin{tabular}{l c c c c c}
    \toprule
     &  \multicolumn{2}{ c }{\textit{Captioning}} & \multicolumn{3}{ c }{\textit{Actions and objects}}\\
     Model & \cider~\cite{Vedantam2014CIDErCI} & \meteor~\cite{banerjee-lavie-2005-meteor} & V-\iou & N-\iou & NC-\iou \\
    \midrule
    Worst & 10.9 & 45.1 & 29.2 & 35.8 & 30.7\\
    Second best & 11.9 & 46.4 & 30.9 & 35.8 & 30.6 \\
    \textbf{Best (Ours)} & \textbf{13.5} & \textbf{48.4} & \textbf{33.7} & \textbf{39.2} & \textbf{32.9}\\
    \bottomrule
  \end{tabular}
  }
  \vspace{-0.2cm}
  \caption{Effect of the rank of our sampled view on the view selection performance on Ego-Exo4D~\cite{grauman2023ego}. Significance, $p \leq 0.05$.}
  \label{tab:predViewRank_autoMetrics_Ego-Exo4D}
\end{table}

Table~\ref{tab:predViewRank_autoMetrics_Ego-Exo4D} shows the impact of the rank of our 
sampled view
on view selection performance.
We observe that the lower the rank of our 
sampled view is, within our model's learned view order, the worse 
our view selection performance is. This shows that our model's learned 
ranking of 
views is highly correlated with the
view quality, which indicates that our model successfully builds an implicit understanding of which views are more informative.

\section{Attention heatmaps of our view selector}\label{sec:supp_attnHeatMaps}

\begin{figure}[t]
    \centering
    \includegraphics[width=1\linewidth]{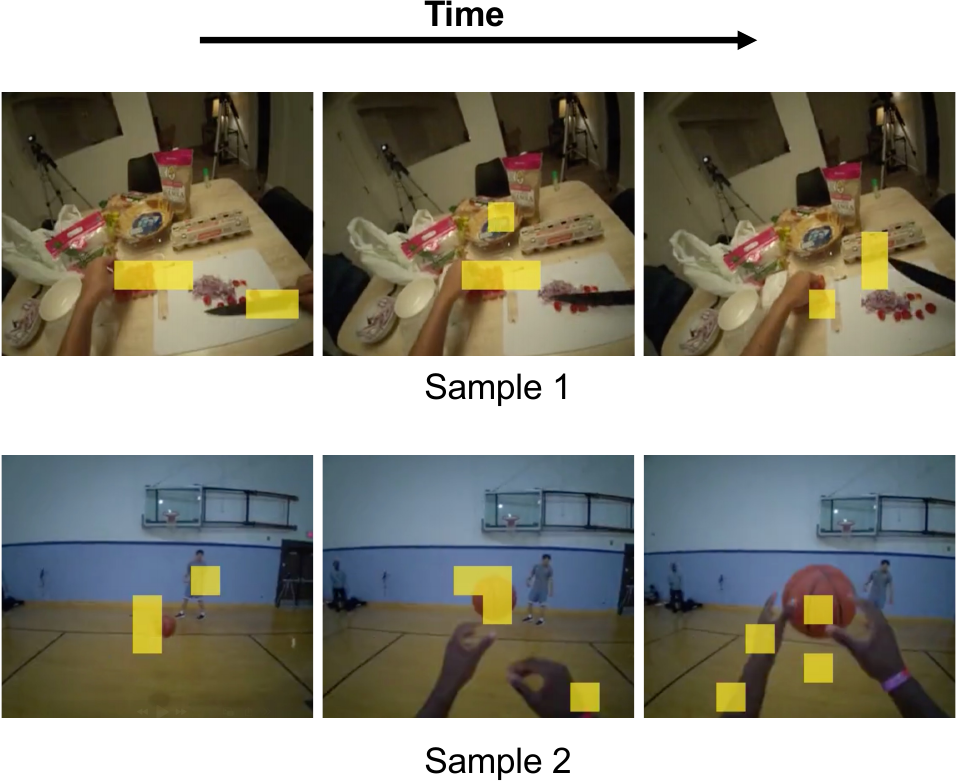}
\caption{\SM{Our model’s attention heatmaps on two best view clips from Ego-Exo4D~\cite{grauman2023ego}. Yellow patches indicate highest attention.}}
\label{fig:attn_maps}
\end{figure}

In Fig.~\ref{fig:attn_maps}, we provide examples of our model's attention heatmaps on Ego-Exo4D~\cite{grauman2023ego}. Our model tends to focus on the salient objects for an activity, even if they are dynamic, indicating its strong activity understanding ability.

\subsection{Analysis of our best view pseudo-labeler}\label{sec:supp+pseudolabelerAnalysis}
\begin{figure*}[t]
    \centering
    \includegraphics[width=1\linewidth]{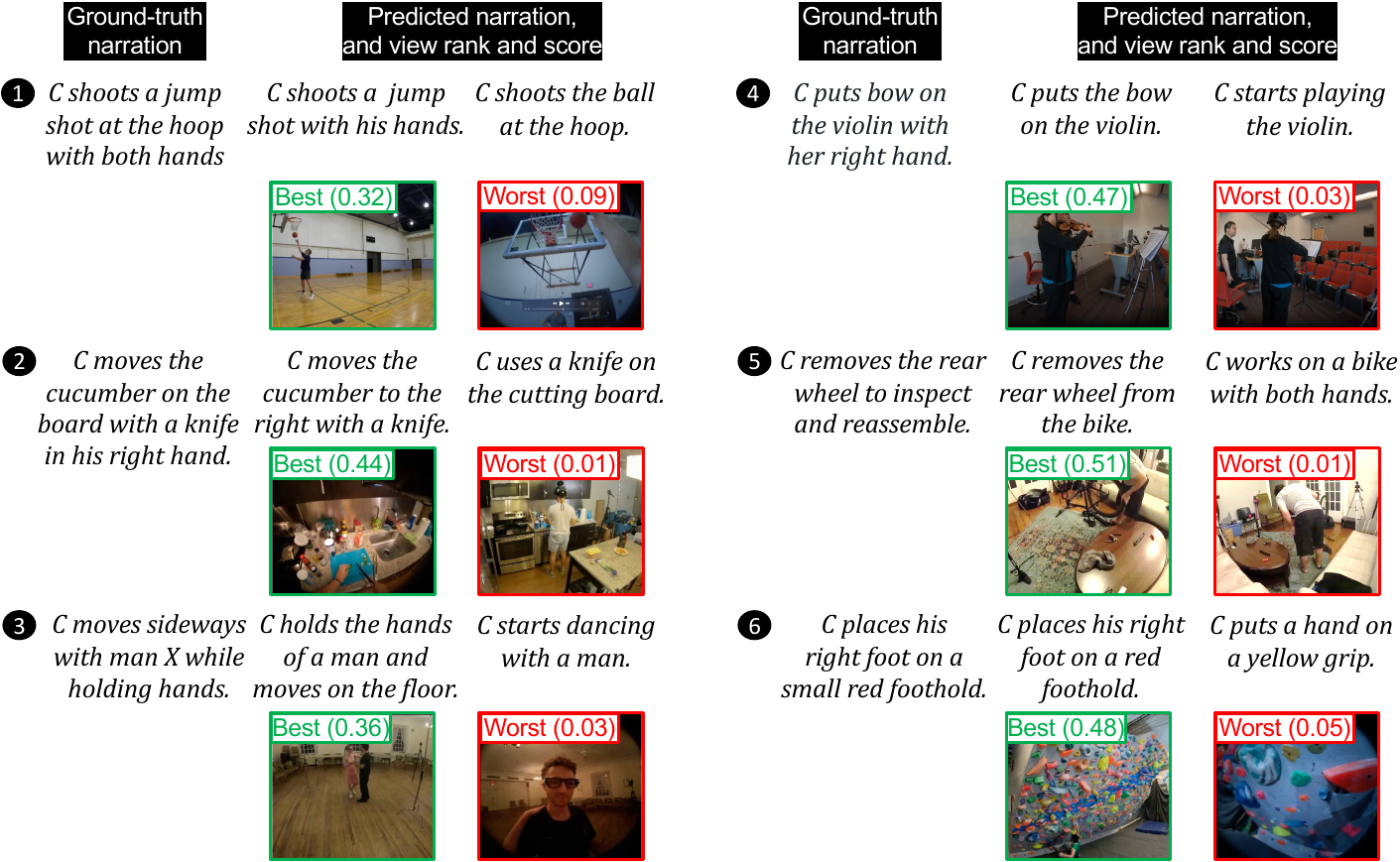}
\caption{Examples of predicted narrations, and the ranks and scores of the views, per our pseudo-labeler $L$, shown alongside ground-truth narrations, in addition to what is provided in
Sec.~\ref{sec:captions} 
in main.}
\label{fig:pseudoLabelExamples_supp}
\end{figure*}

\begin{figure*}[t]
    \centering
    \includegraphics[width=1\linewidth]{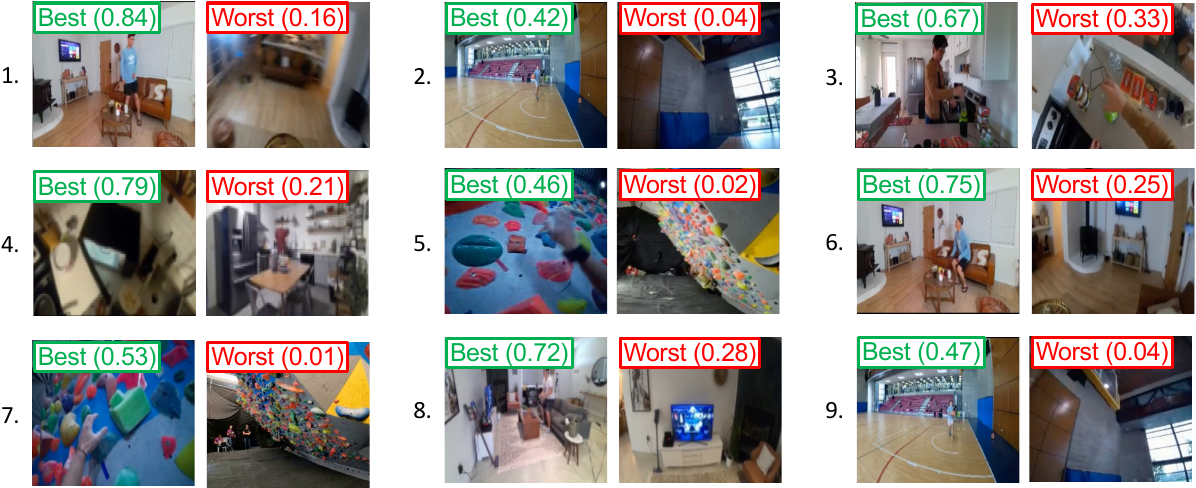}
\caption{Additional examples of best and worst views, and their scores, per our pseudo-labeler $L$.}
\label{fig:pseudoLabelExamples_supp_2}
\end{figure*}

\begin{table}[!t]
\small
  \centering
  \setlength{\tabcolsep}{4pt}
    \begin{tabular}{c c c c c| c c}
    \toprule
    \multicolumn{5}{ c|}{\textit{Ego-Exo4D~\cite{grauman2023ego}}} & \multicolumn{2}{ c }{\textit{LEMMA~\cite{jia2020LEMMA}}}\\
     Ego & Exo 1 & Exo 2 & Exo 3 & Exo 4 & Ego & Exo\\
    \midrule
    \quad 20.4 & 19.8 & 20.3 & 19.6 & 19.9 & 63.6 & 36.4\\
    \bottomrule
  \end{tabular}
  \caption{Probability distribution in \% of our best view pseudo-labels.}
  \label{tab:pseudolabel_probDist}
\end{table}

\begin{table}[!t]
\small
  \centering
  \setlength{\tabcolsep}{4pt}
  \resizebox{1.0\linewidth}{!}{
    \begin{tabular}{l c c c c c c}
    \toprule
     Model & \cider & \meteor  & & V-\iou & N-\iou & NC-\iou\\
    \midrule
    Ours w/ 2 captioners & 13.3 & \textbf{48.4} & & \textbf{34.2} & 38.1 & 32.5 \\
    \textbf{Ours} (w/ 3 captioners) & \textbf{13.5} & \textbf{48.4} &  & 33.7 & \textbf{39.2} & \textbf{32.9}\\
    \bottomrule
  \end{tabular}
  }
  \vspace{-0.2cm}
  \caption{
  \SM{Impact of captioner count on view selection performance, evaluated with Ego-Exo4D~\cite{grauman2023ego}. Significance, $p \leq 0.05$. See row 3 of Table 3, and Sec. 4.2, in main, for results with 1 captioner.}
  }
  \label{tab:ablation_captionerCount}
\end{table}

Here, we analyze different aspects of our pseudo-labeler $L$ (Sec.~\ref{sec:captions} in main).

In table~\ref{tab:pseudolabel_probDist}, we report the distribution of our selected views for both Ego-Exo4D~\cite{grauman2023ego} and LEMMA~\cite{jia2020LEMMA} datasets. For Ego-Exo4D, our model produces a more or less uniform distribution over all views, indicating that depending on the activity and its level of body and object motion, our model can prefer the ego view or one of the exo views with almost equal likelihood. However, for LEMMA, our model tends to prefer the ego view much more than the exo view, re-emphasizing the prevalence of household activities that largely require the ego view for capturing their informative aspects 
(`Dataset' in Sec.~\ref{sec:exp_setup} in main).  

In addition to the ones provided in Fig. 2b in main, we show more pseudo-labeler outputs, comprising view ranks and predicted narrations, alongside the ground-truth narrations, in Fig.~\ref{fig:pseudoLabelExamples_supp}. In Fig.~\ref{fig:pseudoLabelExamples_supp_2}, we provide more such examples without narrations. We see very similar patterns in these additional samples---the better our pseudo-labeler considers a view to be, the more accurate the narration predicted from the view, is, in terms of capturing important activity details. 

In Table~\ref{tab:ablation_captionerCount}, we compare our view selection performance on Ego-Exo4D~\cite{grauman2023ego}, when using 3 vs. 2 captioners---see row 3 of Table 3, and Sec.~4.2, in main for results with 1 captioner, in our pseudo-labeler (Sec.~3.3 in main). Our view selection performance general improves with the increase in the captioner count in our pseudo-labeler, possibly because having more captioners vote on the best view reduces captioning noise and improves pseudo-label quality.

\subsection{Ego-Exo4D with single exo camera}\label{sec:supp_egoExo_singleExo}
\begin{table*}[!t]
\small
  \centering
  \setlength{\tabcolsep}{4pt}
    \begin{tabular}{l c c c c c}
    \toprule
     &  \multicolumn{2}{ c }{\textit{Captioning}} & \multicolumn{3}{ c }{\textit{Actions and objects}}\\
     Model & \cider~\cite{Vedantam2014CIDErCI} & \meteor~\cite{banerjee-lavie-2005-meteor} & V-\iou & N-\iou & NC-\iou\\
    \midrule
    Ego & 10.2 & 45.2 & 30.2 & 34.1 & 29.1\\
    Random & 9.8 & 44.5 & 29.0 & 34.9 & 28.5\\
    Random-exo & 9.6 & 43.8 & 28.0 & 34.2 & 27.4\\
    Hand-object~\cite{cheng2023towards} & 11.5 & 46.8 & 32.2 & 36.8 & 30.5 \\
    Body-area~\cite{Jiang2023RTMPoseRM} & 10.3 & 45.4 & 30.2 & 34.4 & 28.4 \\
    Joint-count~\cite{Jiang2023RTMPoseRM} & 9.9 & 44.6 & 28.6 & 34.1 & 28.1 \\
    Pixel-objectness~\cite{Xiong_2018_ECCV, 10.1145/3183794} & 11.2 & 46.1 & 30.9 & 35.9 & 29.4 \\
    Longest-caption & 0.0 & 0.0 & 0.0 & 0.0 & 0.0 \\
    \textbf{Ours} & \textbf{12.7} & \textbf{47.1} & \textbf{32.7} & \textbf{37.3} & \textbf{30.9} \\
    \bottomrule
  \end{tabular}
  \caption{View selection with Ego-Exo4D, when the candidate viewpoints comprise the 
  ego view and one 
  exo view. All metrics, expressed in $\%$ are averaged over all 
  possible 
  ego-exo view pairs. Significance, $p \leq 0.05$.}
  \label{tab:egoExo_1exo}
\end{table*}

Here, we evaluate our view selector on the single exo camera variant of Ego-Exo4D~\cite{grauman2022ego4d} in order to emulate more typical instructional settings~\cite{miech2019howto100m, jia2020LEMMA} that consist of a single exo camera, but also retain the challenges in the Ego-Exo4D data arising from the diversity in scenarios, varying degrees of body and object motion, etc. Table~\ref{tab:egoExo_1exo} shows the results, where all metrics are first computed separately for each possible ego-exo view pair and then averaged over all pairs. Our model significantly outperforms all baselines across metrics, showing that it is robust to different camera setups even on challenging datasets with diverse activity scenarios and varying levels of motion of the objects and body parts involved in the activity.

\section{3-fold evaluation on Ego-Exo4D}\label{sec:supp_3foldEval}
\begin{table}[!t]
\small
  \centering
  \setlength{\tabcolsep}{4pt}
  \resizebox{0.8\linewidth}{!}{
    \begin{tabular}{l c c c c c c}
    \toprule
     Model & \cider & \meteor & & V-\iou & N-\iou & NC-\iou\\
    \midrule
    Body-area & 10.5 & 46.6 &  & 30.0 & 35.2 & 30.4\\
    \textbf{Ours} & \textbf{11.4} & \textbf{46.9}  & & \textbf{31.2} & \textbf{37.0} & \textbf{31.9}\\
    \bottomrule
  \end{tabular}
  }
  \vspace{-0.2cm}
  \caption{Average view selection results over three disjoint test splits from Ego-Exo4D~\cite{grauman2023ego}. Significance, $p \leq 0.05$.
  }
  \label{tab:multi_testSplits}
\end{table}

In Table~\ref{tab:multi_testSplits}, we report the results from 3-fold evaluation with Ego-Exo4D~\cite{grauman2023ego}. Our model significantly outperforms Body-Area, the best baseline. This shows that our model's improvement over the baselines sustains across multiple test datasets.

\section{Pseudo-labeling cost}\label{sec:supp_pseudoLabelingCost}
We use 8 NVIDIA V100 GPUs for training and performing inference with the captioners in our pseudo-labeler (Sec.~3.2 in main). When pseudo-labeling Ego-Exo4D~\cite{grauman2023ego}, it takes $\sim$2.5 days with VideoLlama captioners, and 3 hours with VideoChat2. For LEMMA~\cite{jia2020LEMMA}, the same takes 1 hour per captioner. Importantly, this is a one-time cost since we pseudo-label only once per dataset, and we do not use any captioner when training or evaluating our view selector.

\section{Model performance vs. distribution of concepts in ground-truth train narrations}\label{sec:supp_perfVsNarrationConceptDist}
\begin{figure*}[!t]
    \centering
    \includegraphics[width=0.5\linewidth]{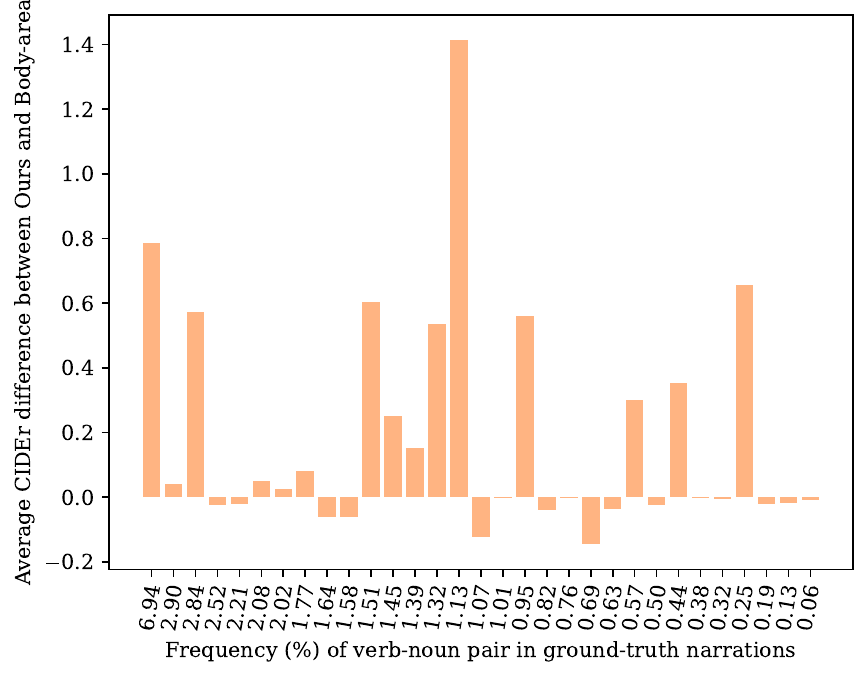}
\caption{\emph{Test} CIDEr difference between our model and the Body-area~\cite{Jiang2023RTMPoseRM} baseline vs. verb-noun pair frequency in \emph{train} narrations, sorted in decreasing order}
\label{fig:supp_perfVsNarrationConceptDist}
\end{figure*}

Fig.~\ref{fig:supp_perfVsNarrationConceptDist} plots our \emph{test} gains over Body-area~\cite{Jiang2023RTMPoseRM}, the strongest baseline, versus the frequency (most to least) of occurrence of different concepts in the ground-truth \emph{train} narrations. The lack of a strong correlation demonstrates that our view selection is not biased by the dominant concepts in the training narrations.

\subsection{Dataset details}\label{sec:supp_dataset}
Here, we provide additional dataset details. For both Ego-Exo4D~\cite{grauman2023ego} and LEMMA~\cite{jia2020LEMMA}, we uniformly sample 8 frames from each clip and resize each frame to $224 \times 224$. Further, we normalize each pixel in a frame by first dividing it by 255 so that its value lies in $[0, 1]$, then subtracting the pixel mean and finally dividing by the pixel standard deviation, where the pixel mean and standard deviation are channel-specific. We set the mean and standard deviation to $[0.48145466, 0.4578275, 0.40821073]$ and $[0.26862954, 0.26130258, 0.27577711]$, respectively, for our view selector and Video-Llama~\cite{zhang2023video} captioners, and $[0.485, 0.456, 0.406]$ and $[0.229, 0.224, 0.225]$, respectively, for our VideoChat2~\cite{li2023mvbench} captioner, where the channels follow the RGB order. 

We split the Ego-Exo4D videos into sequences of clips, each coupled with a narration,
by adopting the ``contextual variable length clip pairing strategy" strategy~\citep{lin2022egocentric, ramakrishnan2023naq}, which
generates
temporal windows for 
extracting clip-narration pairs. 
To split the LEMMA videos into clips,
we group contiguous frames using their verb and noun annotations 
(Sec.~\ref{sec:exp_setup} in main).

For Ego-Exo4D, we preprocess each narration by denoting each activity participant mentioned in the narration using `X$i$', where $i$ is the participant's position in the sequence in which the participants appear in the time-sorted narrations for each full video (a take in Ego-Exo4D). The value of $i$ starts from 0. We produce narrations for LEMMA by appending the verb and object annotations, where each narration has the following structure: `verb1: object1\_1, object1\_2, ...; verb2: object2\_1, object2\_2, ...; ...' .

\subsection{Implementation details}\label{sec:supp_implementation}
Here, we provide additional implementation details for different components of our framework, and our Pixel-objectness~\cite{Xiong_2018_ECCV, 10.1145/3183794} baseline.

\subsubsection{Captioner}
For our VideoLlama~\cite{zhang2023video} and VideoChat2~\cite{li2023mvbench} captioners, we use a model with the same architecture as proposed in the original paper and initialize the parameters from the checkpoints released by the authors.
We freeze the ViT~\cite{dosovitskiy2020image} encoder and LLM (without LoRA~\cite{hu2021lora}, wherever it is used) in all captioners, and train all other modules with an AdamW~\cite{loshchilov2017decoupled} optimizer for a maximum of 1.6 million iterations. We use a cosine annealing learning rate schedule~\cite{loshchilov2016sgdr} with a linear warmup over 5000 iterations, where we set the starting learning rate to $10^{-6}$, the peak learning rate to $3 \times 10^{-5}$, and the minimum learning rate during cosine annealing to $1 \times 10^{-5}$.  We set the total batch size to 8, and the $(\beta_1, \beta_2)$ and weight decay in AdamW to $(0.9, 0.999)$ and $5 \times 10^{-2}$, respectively. Furthermore, for VideoChat2, we turn off flash attention~\cite{dao2022flashattention, dao2023flashattention}. Finally, we set the LLM prompt to `What is the person wearing smart glasses doing in the video?' for Ego-Exo4D~\cite{grauman2023ego} and `What is the person wearing a head-mounted camera in the video doing?' for LEMMA~\cite{jia2020LEMMA}. 

\subsubsection{View selector}
We use the EgoVLPv2~\cite{pramanick2023egovlpv2} vision encoder, pretrained on Ego-Exo4D~\cite{grauman2023ego}, to obtain visual features $f$ in our view selector $S$ (Sec.~\ref{sec:selector} in main). The EgoVLPv2 encoder is a 12-layer TimeSformer~\cite{bertasius2021space} model, where we set the prediction head ($head$), prediction logits ($pre\_logits$) and fully-connected layer ($fc$) to identity functions from PyTorch. We attach a shared convolution layer to the encoder for producing shared features for both view classification in $W$ (Sec.~\ref{sec:exp_setup} in main) and pose prediction in $P$ (Sec.~\ref{sec:exp_setup} in main). The shared convolution has a kernel size, padding and stride of 1, 768 input channels and 192 output channels. The output of the shared convolution goes into a view selection head and a pose prediction head. 

The view selection head begins with the following layers: 1) a Batch Normalization~\cite{10.5555/3045118.3045167} layer with 192 input channels, 2) a ReLU~\cite{agarap2018learning} activation, 3) a convolution layer with a kernel size of 4, stride of 2, padding of 1, and 192 and 96 input and output channels, respectively, 4) a Batch Normalization layer with 96 input channels, 5) a ReLU activation, and 6) a convolution layer with a kernel size of 4, stride of 2, padding of 0, and 96 and 24 input and output channels, respectively. We feed the output of the last convolution from above to a a transformer~\cite{vaswani2017attention} encoder, which comprises 2 layers with 8 heads and 768 channels. Each layer uses a dropout of 0.1 and uses sinusoidal positional encodings~\cite{vaswani2017attention}. We then feed the output of the transformer encoder to a 2-layer MLP that comprises 1) a linear layer with 768 input channels and 128 output channels, 2) a Batch Normalization layer with 128 input channels, 3) a ReLU activation, 4) a dropout layer with the dropout probability set to 0.1, and 5) a linear layer with 128 input channels and the output channel count set to the number of views. 

The pose prediction head comprises a convolution-only and linear-layer-only component. The convolution-only component comprises 1) a Batch Normalization~\cite{10.5555/3045118.3045167} layer with 192 $\times$ 2 = 384 input channels, 2) a ReLU~\cite{agarap2018learning} activation, 3) a dropout layer with the dropout probability set to 0.1, and 4) a convolution layer with a kernel size of 4, stride of 2, padding of 1, and 384 and 48 input and output channels, respectively. The linear-layer-only component is comprised of 1) a Batch Normalization layer with 2352 input channels, 2) a ReLU activation, 3) a dropout layer with the dropout probability set to 0.1, 3) a linear layer with 2352 input dimensions and 53 output dimensions. We feed the outputs of the convolution-only component to the linear-layer-only component.

We employ $resize$ and $reshape$ operations from PyTorch wherever necessary. 

We train our view selector using AdamW~\cite{loshchilov2017decoupled} with a learning rate of $10^{-5}$ for the EgoVLPv2~\cite{pramanick2023egovlpv2} vision encoder and $10^{-4}$ for the rest of the model. We set the total batch size to 24, and the $(\beta_1, \beta_2)$ and weight decay in AdamW to $(0.9, 0.999)$ and $10^{-5}$, respectively.

For all our model components, we stop training once the validation loss starts increasing. 

\subsubsection{Baseline: Snap angles~\cite{Xiong_2018_ECCV, 10.1145/3183794}}
This baseline
(`Baselines' in
Sec.~\ref{sec:exp_setup} in main) 
is an upgrade to the most relevant existing methods~\cite{Xiong_2018_ECCV, 10.1145/3183794} in the literature.
It predicts the view with the highest count of pixels belonging to foreground~\cite{Xiong_2018_ECCV, 10.1145/3183794} and salient~\cite{10.1145/3183794} objects but not lying near the frame boundaries~\cite{Xiong_2018_ECCV}, as the best view. To do so, we treat the set of all objects mentioned in the training narrations as foreground and salient, and query a model composed of GroundingDino~\cite{liu2023grounding} and Segment Anything (SAM)~\cite{kirillov2023segment} with this set to detect its constituent pixels. Specifically, we first feed GroundingDino with the foreground-and-salient object set to compute the corresponding bounding boxes. Next, we feed these bounding boxes to SAM to mark all pixels of relevance. Finally, for each view, we compute a score that is a weighted sum of its average foreground-and-salient pixel count across all frames and a penalty term that lowers the count by the inverse of the view's frame count, for every pixel found within a certain distance from the frame boundaries. We set the weights on the foreground-and-salient pixel count to 1.0, and the penalty term to 0.1 and 0.02 for Ego-Exo4D~\cite{grauman2023ego} and LEMMA~\cite{jia2020LEMMA}, respectively, through validation, and the distance for using a foreground-and-salient pixel in computing the penalty term, to $6.25\%$~\cite{Xiong_2018_ECCV} of the frame size.

\end{document}